%% file: main.tex
\documentclass{article}
\usepackage{iclr2027_conference,times}

\input{math_commands.tex}

\usepackage[utf8]{inputenc}
\usepackage[T1]{fontenc}
\usepackage[strings]{underscore}
\usepackage{amsmath,amsfonts,amssymb}
\usepackage{algorithm}
\usepackage{algpseudocode}
\usepackage{booktabs}
\usepackage{graphicx}
\usepackage{listings}
\usepackage{microtype}
\usepackage{multirow}
\usepackage{nicefrac}
\usepackage{pifont}
\usepackage{subcaption}
\usepackage{tabularx}
\usepackage{adjustbox}
\usepackage{xcolor}
\usepackage{tcolorbox}
\usepackage{wrapfig}
\usepackage{xspace}
\usepackage[hidelinks]{hyperref}
\usepackage{url}

\newtcolorbox{findingbox}{
  colback=gray!10, colframe=black!80,
  boxrule=1pt, arc=4pt,
  left=6pt, right=6pt, top=6pt, bottom=6pt,
  before skip=6pt, after skip=6pt,
}

\newcommand{\system}{{\textsc{NovelAPIBench}}\xspace}
\newcommand{\writingTODO}[1]{\def\tmp{#1}\textcolor{purple}{\bf [TODO]\ifx\tmp\empty\else: #1\fi}}

\title{\system: Diagnosing How A Code LLM Learns to Use Novel APIs}

\author{%
\textbf{Jinnuo Liu}$^{1,2}$\quad
\textbf{Yue Peng}$^{2,*}$\quad
\textbf{Jinhan Niu}$^{2,*}$\quad
\textbf{Hongyi Wen}$^{1,2,\dagger}$\\[6pt]
$^{1}$New York University\\
$^{2}$Center for Data Science, NYU Shanghai\\[6pt]
{\small $^{*}$Equal contribution\quad $^{\dagger}$Corresponding author}\\[4pt]%
}

\iclrfinalcopy

\begin{document}

\maketitle
\lhead{Preprint}

\begin{abstract}
  \input{sec/0_abstract}

  \label{sec:abstract}
\end{abstract}

\section{Introduction}
\label{sec:intro}
\input{sec/intro}

\section{Related Work}
\label{sec:related_work}
\input{sec/related_work}

\section{\system}
\label{sec:method}
\input{sec/method}

\section{Experiments}
\label{sec:experiments}
\input{sec/experiments}

\section{Discussions}
\label{sec:discussions}

\input{sec/discussions}

\section{Conclusion}
\label{sec:conclusion}
\input{sec/conclusion}

\clearpage
\bibliography{references}
\bibliographystyle{iclr2027_conference}

\clearpage
\appendix

\input{app/a_limitations}
\input{app/b_benchmark}
\input{app/c_failtax}
\input{app/d_setup}

\input{app/e_rq1}
\input{app/f_rq2}
\input{app/g_rq3}

\input{app/i_dataset_doc}

\end{document}

%% file: math_commands.tex
\usepackage{amsmath,amsfonts,bm}

\def\eqref#1{equation~\ref{#1}}

\def\1{\bm{1}}

\DeclareMathAlphabet{\mathsfit}{\encodingdefault}{\sfdefault}{m}{sl}
\SetMathAlphabet{\mathsfit}{bold}{\encodingdefault}{\sfdefault}{bx}{n}

%% file: sec/0_abstract.tex
\input{sec/abstract}

%% file: sec/abstract.tex
Rapidly evolving software libraries require coding language models to use unfamiliar APIs. Yet failed solutions alone tell us little about what API knowledge models need or how to provide it. We present \system{}, an automated diagnostic framework that builds benchmarks around the API knowledge gaps of a target model. It identifies novel APIs, generates executable coding tasks with separately controllable knowledge components, and classifies failures into six categories. The framework can be rerun as models and libraries evolve. We study 21 libraries across five domains and six backbone models, with our primary benchmark comprising 1,670 tasks covering 856 APIs. Our experiments show that when supplied directly, usage examples provide the strongest standalone guidance for novel API acquisition, and combining them with signature descriptions approaches full-knowledge performance. Retrieving this knowledge from a pool of APIs yields lower accuracy than supplying only the target API's knowledge, even when retrieval finds the target. Fine-tuning on other APIs improves performance on unseen APIs mainly when external knowledge is supplied, helping models both select the target API and integrate it into the surrounding code. These findings highlight complementary roles for external API knowledge and the ability to apply it. By separating knowledge content, delivery, and use, \system{} helps diagnose what limits novel API use and guides adaptation to evolving libraries. Code and data are available at \url{https://github.com/MAPS-research/NovelAPIBench}.

%% file: sec/intro.tex
Software libraries evolve quickly. New releases add functions, change interfaces, and reorganize modules, leaving coding models to work with APIs that may not be covered by their training data. When a model fails on a new API, the failure itself gives little guidance on how to adapt it. Should we provide the API signature, retrieve a usage example, expose more detailed documentation, or update the model parameters? This raises a basic question for adapting coding models to evolving libraries: \emph{what API knowledge is the model missing, and how should that knowledge be provided?}

Answering this question requires separating \textbf{knowledge content} from \textbf{knowledge delivery}. API information may come from signatures, examples, semantic descriptions, or implementation details, while that information may be supplied through retrieval or incorporated through parametric adaptation~\citep{lewis2020rag, gao2023ragsurvey, hu2022lora, rome, memit, grace, alphaedit}. Different knowledge sources may address different gaps, and their usefulness may also depend on how the information is integrated into the model. 

Existing API-evolution benchmarks provide limited support for diagnosing such gaps~\citep{codeupdatearena, versicode, libevolutioneval, gitchameleon, rustevo}. API novelty is model-dependent, so a fixed benchmark becomes less informative as newer models absorb more recent code. At the same time, aggregate metrics such as $\mathrm{pass}@k$ reveal whether a solution succeeds but give little insight into what knowledge was missing. Studying novel API acquisition therefore requires an evaluation that adapts to the target model, controls what API information it receives, and explains failures beyond a single correctness score.

\begin{figure}[t]
    \centering
    \includegraphics[width=0.9\linewidth]{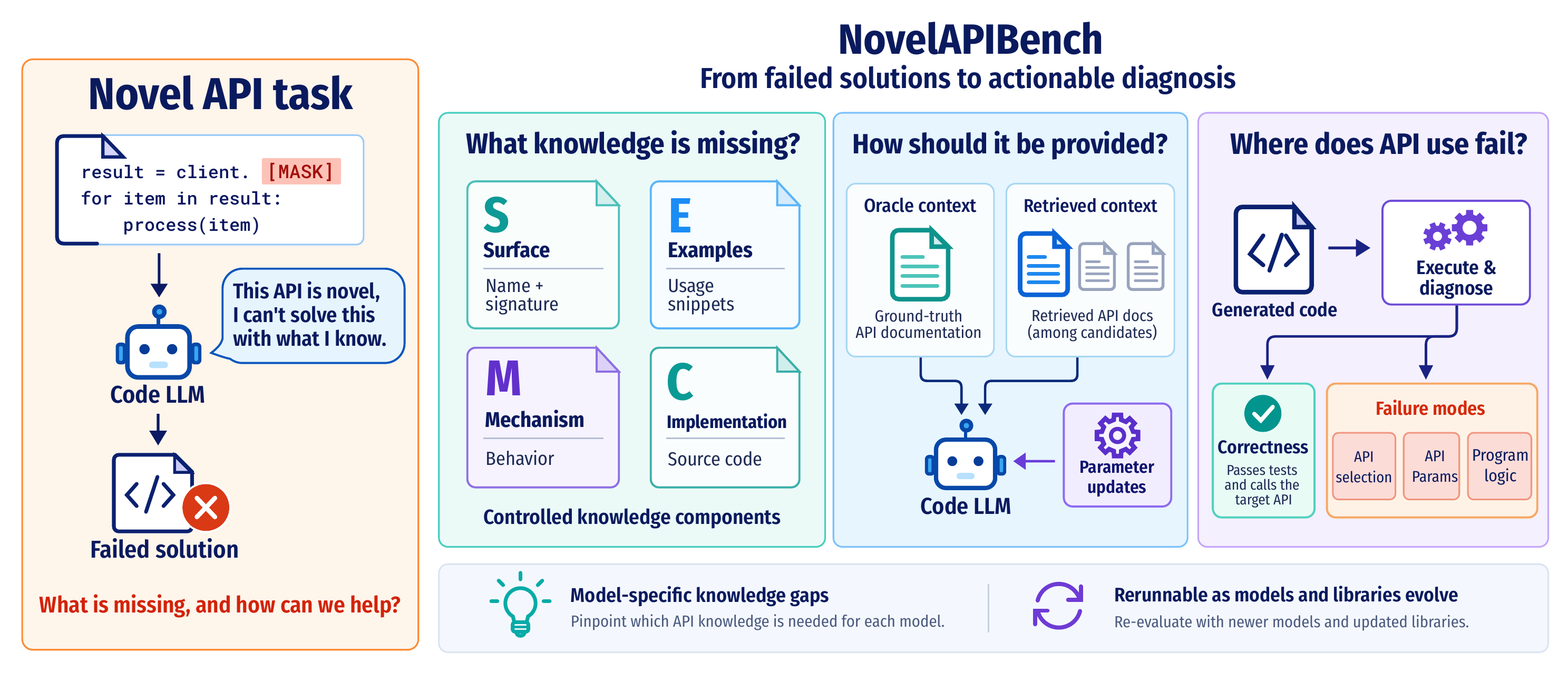}
    \caption{Overview of \system{}. Starting from novel API tasks a target model cannot solve without external knowledge, the framework investigates what API knowledge helps, how it should be provided, and where API use fails. }
    \label{fig:nab_headline}
\end{figure}

We therefore introduce \system{}, a model-conditioned diagnostic framework for novel API acquisition. Given a target model and library, \system{} discovers recently introduced APIs and filters for cases that expose an empirical knowledge gap in that model. It then constructs executable coding tasks together with \textbf{decomposed API knowledge} that can be selectively provided to the model. Each prediction is executed in its program context by an automated harness, which checks whether the model correctly calls the target API as the reference solution does and records the failure types of unsuccessful solutions. This design makes the benchmark \textbf{evolvable}: as models and libraries change, the same pipeline can be rerun to identify the knowledge gaps that remain relevant. It also makes the evaluation \textbf{actionable}: by linking controlled knowledge interventions to concrete failure modes, \system{} helps identify what information a model needs and how that information should be delivered. We apply the pipeline to 21 libraries across five domains and evaluate six backbone models with model-conditioned benchmark instances. Our primary Qwen2.5-Coder-7B evaluation covers 856 APIs and 1,670 executable tasks from 20 libraries.

To isolate the value of different knowledge components, we first supply models directly with information about the target API. Across six backbones, \textbf{usage examples provide the strongest standalone guidance}, and combining them with signature descriptions approaches full-knowledge-bundle performance. Explicit interface information is particularly useful for APIs whose signatures have changed, where examples alone may not reveal the updated requirements. These results show how the knowledge needed for effective API use depends on the type of API novelty.

We next evaluate model performance when API knowledge must be obtained through retrieval rather than provided directly. Retrieving from a pool of candidate APIs greatly lowers accuracy compared with directly supplying the target API's knowledge, although examples remain the most useful knowledge component. Crucially, \textbf{correctly retrieving the target API does not close the performance gap}: losses persist even on tasks where retrieval succeeds. Effective knowledge delivery therefore requires more than simply identifying the correct API; it also depends on how the retrieved knowledge is presented and used.

Finally, \textbf{fine-tuning improves the use of external knowledge on unseen APIs}. SFT and RAFT trained on other APIs substantially improve performance when the target API's knowledge is supplied, while offering little gain without it. Our failure analysis shows improvements both in selecting the target API and in integrating the call into the surrounding code. By contrast, the tested knowledge-editing methods perform poorly on Novel APIs without supplied knowledge and yield no significant gains over the base model when that knowledge is provided.

Overall, these findings suggest complementary roles for external knowledge and parameter updates: providing evolving API details and improving the ability to apply them. By distinguishing these needs, \system{} helps identify what still limits novel API acquisition and use.

Our contributions are:
\begin{itemize}
    \item We introduce \system{}, a model-conditioned, execution-based framework that constructs diagnostic benchmarks around the API knowledge gaps of a target coding model and can be rerun as models and libraries evolve.
    
    \item We develop a controlled evaluation protocol that separates API knowledge content from delivery and connects knowledge interventions to six categories of execution failures.
    
    \item We show that examples provide strong guidance, retrieval leaves a performance gap even when the target API is retrieved successfully, and fine-tuning improves the application of supplied knowledge to novel APIs.
\end{itemize}

%% file: sec/related_work.tex
\paragraph{Benchmarks for Evolving APIs.}
Recent benchmarks study API evolution from several angles, including synthetic updates~\citep{codeupdatearena}, version-controllable or temporally stratified evaluation~\citep{versicode,evocodebench,libevolutioneval,cipollone2026libevobench}, compatibility with outdated APIs~\citep{gitchameleon,ashik2026lagbehind,lin2026incompatible}, and automated benchmark construction~\citep{rustevo}. Related work also probes model-specific knowledge boundaries to guide intervention~\citep{wu2026powercodebench}, while error taxonomies have been developed for general code generation and infrastructure-as-code~\citep{wang2025codeerrors,nekrasov2024iac}. \system{} brings these ideas together by constructing tasks around the knowledge gaps of a target model, varying the API information it receives, and tracing how those interventions change concrete failure modes.

\paragraph{Knowledge Decomposition for Code Generation.}
Prior work shows that the form of API documentation matters for code generation. Examples often provide strong guidance~\citep{chen2025docs}, while signatures, source code, and documentation can have different effects depending on the API and retrieval setting~\citep{codenav,jain2024dag, hsieh2023tool}. Recent systems further retrieve API specifications or accumulate usage guidance for unfamiliar and private APIs~\citep{maninger2026webapi,li2026memcoder}. We study these information sources under controlled conditions, separating signatures, examples, semantic descriptions, and implementation source and relating each intervention to execution failures.

\paragraph{Parametric Adaptation and Knowledge Editing.}
Novel software knowledge can also be incorporated through parameter updates. Prior work studies model editing for code~\citep{grace,rome,memit,alphaedit,lmscode-editing,robust-ke-code}, supervised adaptation to changing or private APIs~\citep{codesync,jiang2026koco,zhang2026pricoder}, and broader comparisons between fine-tuning, retrieval, and knowledge injection~\citep{bohnet2025sft-lora-icl,kginjection-selfdistil}. CURE~\citep{tran2026cure} additionally targets deprecated API behavior through contrastive unlearning. We compare retrieval, fine-tuning, and editing within the same benchmark and knowledge decomposition, allowing us to examine whether these methods supply API-specific information or improve how models use that information.

%% file: sec/method.tex
\subsection{Problem Formulation}

We study novel API acquisition relative to a target model $\mathcal{M}$ and a Python library $\mathcal{L}$. We first identify APIs introduced or modified after a specified release boundary, then select coding tasks that $\mathcal{M}$ fails to solve without additional API information under our sampling protocol. We use this behavioral criterion to operationalize an empirical API knowledge gap. The release boundary defines the candidate API pool, while task selection makes the benchmark model-conditioned.

For each target API $a$, we construct a \textbf{knowledge bundle} with four fields:
\begin{itemize}
    \item \textbf{Surface information}: the fully qualified API name $S_{\text{name}}$ and parameter signature $S_{\text{param}}$, including types, defaults, and parameter descriptions;
    \item \textbf{Examples} $E$: documentation-grounded, LLM-generated usage examples;
    \item \textbf{Mechanism description} $M$: a natural-language explanation of the API's behavior or underlying design;
    \item \textbf{Implementation Code} $C$: the API's implementation source when available.
\end{itemize}
These fields form the controlled knowledge interventions used throughout our experiments.

Each benchmark task centers on one target API and asks the model to complete a masked region of executable Python code. A task includes the natural-language specification, surrounding code, a hidden reference solution, and an executable test harness. It may use other helper APIs, but only one API is designated as the target. This lets us vary the knowledge provided for that API and diagnose failures with a clear target. Full schemas and examples are provided in Appendix~\ref{app:bench-components}.

\begin{figure}[t]
    \centering
    \includegraphics[width=0.95\linewidth]{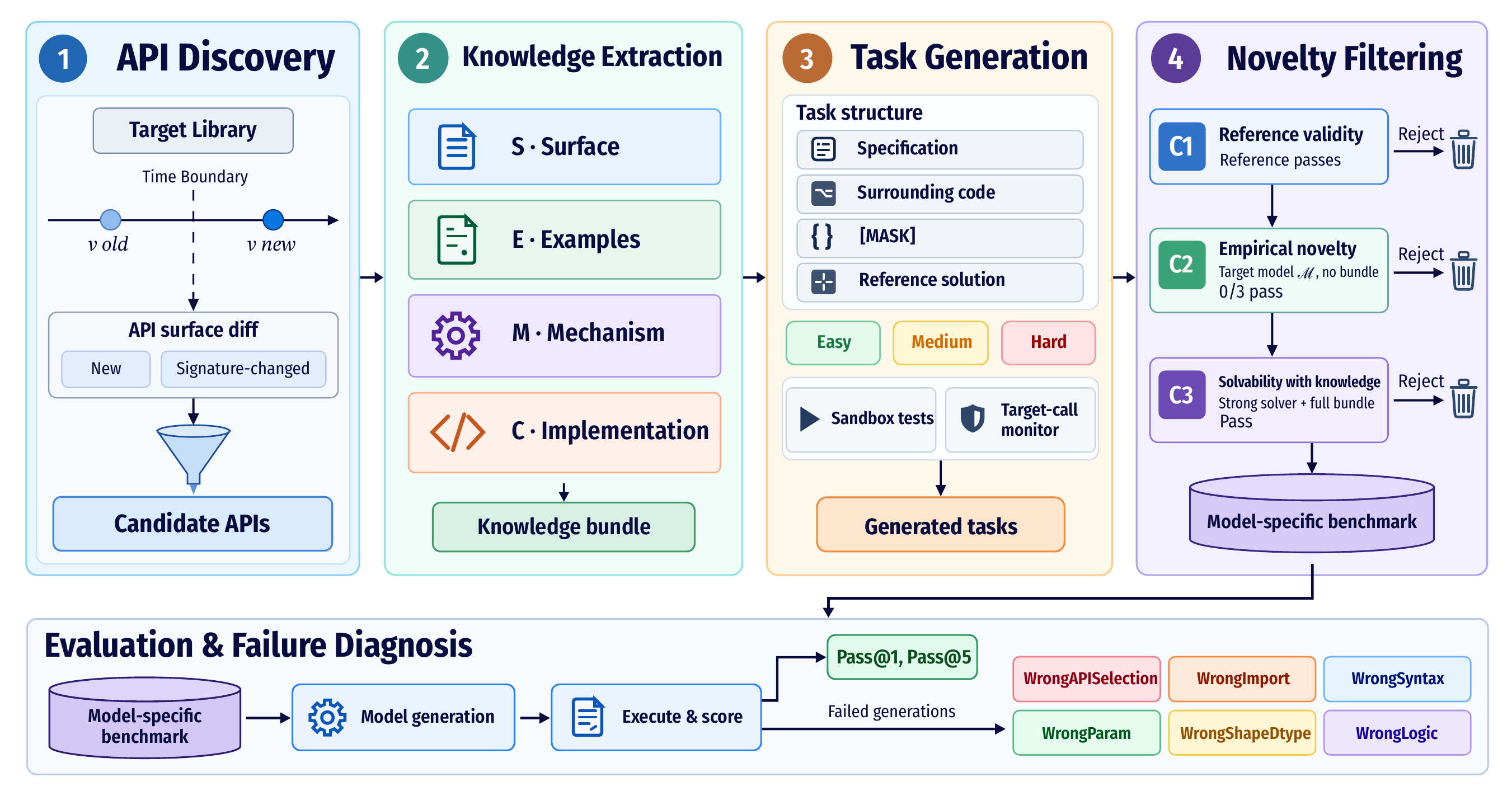}
    \caption{Overview of the \system pipeline. }
    \label{fig:main_fig}
\end{figure}

\subsection{Benchmark Pipeline}

\system{} constructs a model-conditioned benchmark in four stages: API discovery, knowledge extraction, task generation, and quality filtering. Stages~1--3 construct candidate APIs, knowledge bundles, and executable tasks. Stage~4 checks task validity and solvability and applies a separate novelty filter for each target model. Figure~\ref{fig:main_fig} gives an overview.

\paragraph{Stage 1: API Discovery.}
We use a configurable release boundary to identify newly introduced or modified APIs. In our experiments, all backbones share one candidate pool built with December 2023 as the primary boundary (Appendix~\ref{app:bench-pipeline}). For each library, we identify the latest release $v_{\text{old}}$ on or before this boundary and a newer release $v_{\text{new}}$, install both versions, and compare their public API surfaces using Python introspection and AST analysis. Newly introduced or signature-changed callables become candidate APIs. We remove entries that are private, deprecated, inherited from external modules, or unsuitable for automatic task construction. Library versions and filtering rules are provided in Appendix~\ref{app:bench-pipeline}. 

\paragraph{Stage 2: Knowledge Extraction.}
For each candidate API, we build the knowledge bundle from library introspection, official documentation, public usage examples, and source code. Surface information is extracted directly from the installed package, with missing parameter descriptions completed from documentation. Usage examples are generated from the API's documentation and signature, and checked in the sandbox when executable. Mechanism descriptions are grounded in the API's documentation, source, or associated papers. Implementation code is extracted from the installed package with docstrings removed. APIs for which the required bundle fields cannot be constructed are discarded.

\paragraph{Stage 3: Task Generation.}\label{sec:stage3}
For each knowledge bundle, we generate executable coding tasks centered on the target API. Each task contains a natural-language specification, surrounding code, a masked completion, and a reference solution. We execute the reference solution in a sandbox and use its behavior to construct an automated test harness. A target-call monitor verifies that a passing solution actually invokes the target API and reproduces the reference solution's calls to it, so the harness evaluates API use within the program rather than only whether the code runs.

The generator also assigns each task to one of three structural difficulty tiers, with increasing program context and integration around the target API. We validate these tiers using program-structure statistics and human ordering judgments in Appendix~\ref{app:difficulty-validation}. For APIs whose examples cannot be executed independently, we generate three easy variants; these tasks remain subject to the same execution and novelty filters. Full generation details and fallback statistics are provided in Appendix~\ref{app:bench-pipeline}.

\paragraph{Stage 4: Quality and Novelty Filtering.}
\label{sec:stage4}
We retain tasks that pass three checks:
\begin{itemize}
    \item \textbf{C1: Reference validity.}
    The reference solution must pass the executable test harness.

    \item \textbf{C2: Empirical novelty.}
    For each target model $\mathcal{M}$, we sample three solutions at $T=0.8$ without the target API's knowledge bundle. We retain the task for that model only if all three solutions fail. This produces a model-specific task set from the shared candidate pool.

    \item \textbf{C3: Solvability with knowledge.}
    A strong solver receives the task with the Full knowledge bundle and must produce a solution that passes the harness.
\end{itemize}

The pipeline can be updated as models and libraries evolve: applying C2 to a new model selects its task set, while rerunning the full pipeline with a later release boundary and newer library versions refreshes the candidate pool. Appendix~\ref{app:bench-pipeline} reports the number of tasks retained at each stage.

\subsection{Failure Diagnosis}
\label{sec:failtax}

For every unsuccessful generation, we identify the stage at which API use breaks down. We use six mutually exclusive labels:
\texttt{WrongAPISelection} for choosing or hallucinating the wrong API, or bypassing the correct API;
\texttt{WrongImport} for resolving the correct symbol through an incorrect or missing module path;
\texttt{WrongSyntax} for syntactically invalid Python;
\texttt{WrongParam} for incorrect arguments to the target API;
\texttt{WrongShapeDtype} for violations of the API's input contract; and
\texttt{WrongLogic} for errors in the surrounding code when the target API is invoked with acceptable arguments, such as a crash elsewhere in the program, a callback with the wrong contract, or a call whose inputs were prepared incorrectly.

The labels are assigned to individual model generations. For the main $\mathrm{pass}@1$ analysis, we diagnose the greedy output for each task. Simple cases such as syntax errors and unresolved target symbols are classified deterministically. The remaining failures are judged by GPT-5-mini using the task, generated code, extracted call structure, and execution traceback. We validate these labels against independent human annotations in Appendix~\ref{app:human-eval-failtax}.

%% file: sec/experiments.tex
We study novel API use through three questions: \textbf{RQ1:} What knowledge enables effective API use? \textbf{RQ2:} What limits the effectiveness of retrieved knowledge? \textbf{RQ3:} What does parametric adaptation contribute to knowledge acquisition and use?

\subsection{Experimental Setup}
\label{sec:experimental-setup}

We use Qwen2.5-Coder-7B-Instruct~\citep{qwen} as the primary backbone, with $1{,}670$ tasks targeting $856$ novel APIs from $20$ Python libraries across five domains. We replicate the knowledge-component study on five additional backbones (R1-Distill-Qwen-7B~\citep{guo2025deepseek}, Seed-Coder-8B~\citep{seedcoder2025}, OpenCoder-8B~\citep{huang2024opencoder}, Qwen2.5-Coder-14B, Qwen2.5-Coder-32B~\citep{qwen}), each with its own model-conditioned benchmark instance, and use shared task sets for cross-model comparisons. For RQ1, we supply selected components of the target API's knowledge bundle directly in the prompt; for RQ2, we replace this oracle context with the top-five contexts retrieved by BGE-small~\citep{bge}. For RQ3, we compare SFT, RAFT~\citep{zhang2024raft}, GRACE~\citep{grace}, MEMIT~\citep{memit}, and a LoRA-based variant of AlphaEdit~\citep{alphaedit} under an $80/20$ split by target API, evaluating the base and adapted models on $348$ held-out tasks both with and without the full knowledge bundle.

Our primary metric is execution-based $\mathrm{pass}@1$ under greedy decoding.
A completion passes if it executes without error in the task's program context, invokes the target API, and satisfies the reference-based checks of API use (Appendix~\ref{app:bench-task}). We categorize failures using the taxonomy in Section~\ref{sec:failtax} and report $95\%$ confidence intervals from bootstrap resampling clustered by target API. Benchmark statistics and experimental settings are provided in Appendices~\ref{app:bench-stats} and~\ref{app:setup}.

\begin{figure}[t]
    \centering
    \includegraphics[width=\linewidth]{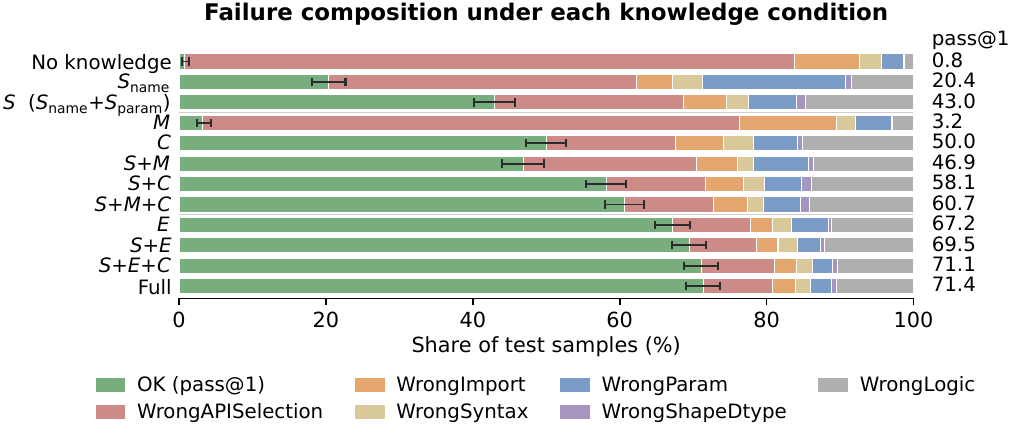}
    \caption{Knowledge-component effects on Qwen2.5-Coder-7B ($1{,}670$ tasks). Bars show success and failure rates; whiskers indicate $95\%$ API-cluster bootstrap intervals for pass@1.}
    \label{fig:rq1-components}
\end{figure}

\subsection{RQ1: What Knowledge Enables Effective API Use?}
\label{sec:rq1}

We first isolate knowledge content by directly supplying the target API's information. We use $S=S_{\text{name}}+S_{\text{param}}$ for surface information, $E$ for examples, $M$ for mechanism descriptions, and $C$ for implementation code; $\textsc{Full}=S+E+M+C$.

\paragraph{Concrete usage examples offer the most effective guidance.}
Examples alone achieve $67.2\%$ pass@1, substantially outperforming mechanism descriptions ($3.2\%$), implementation code ($50.0\%$), and surface information ($43.0\%$; Figure~\ref{fig:rq1-components}). This suggests that showing how an API is used provides more effective standalone guidance than describing its behavior or exposing its implementation. Examples also account for much of the performance achieved with the full bundle: combining examples with surface information raises pass@1 to $69.5\%$, and Full reaches $71.4\%$.

The failure breakdown helps explain this advantage. Providing the API name reduces selection errors from $83.1\%$ to $41.9\%$ of tasks; further adding parameter information reduces argument errors from $19.4\%$ to $6.6\%$. Examples reduce both errors further, suggesting that concrete usage conveys both which API to call and how to call it. 

\begin{figure}[t]
    \centering
    \includegraphics[width=0.95\linewidth]{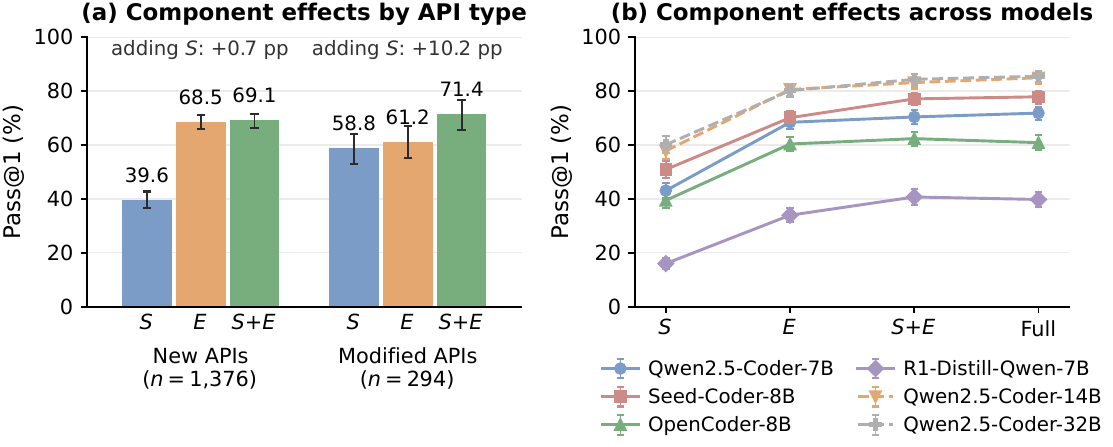}
    \caption{When knowledge components help. (a) On Qwen2.5-Coder-7B, adding $S$ to $E$ changes pass@1 by $+0.7$ pp for newly introduced APIs but $+10.2$ pp for signature-modified APIs. (b) On the $1{,}405$ tasks shared by all six backbones, examples consistently outperform surface information; Full adds little over $S+E$. Whiskers show $95\%$ API-cluster bootstrap intervals. }
    \label{fig:rq1-novelty-backbones}
\end{figure}

\paragraph{Surface information is more useful when an existing API changes.}
Figure~\ref{fig:rq1-novelty-backbones}a shows the decomposition of two types of APIs. The benefit of combining surface information with examples depends on the type of API novelty. For \textbf{newly introduced APIs}, adding $S$ to $E$ leaves performance nearly unchanged ($68.5\%$ to $69.1\%$). However, for \textbf{signature-modified APIs}, it raises pass@1 from $61.2\%$ to $71.4\%$. This difference in gains holds across all six backbones (Holm-adjusted $p\leq0.009$, see Appendix~\ref{app:novelty-type} for details), suggesting that explicit interface information helps clarify changes that usage examples alone may not capture.

\paragraph{The main trends persist across backbones.}
On the common task set, $E$ exceeds $S$ by $17.9$--$25.3$ pp across all six models, while Full differs from $S+E$ by only $-1.5$ to $+1.8$ pp (Figure~\ref{fig:rq1-novelty-backbones}b). Larger Qwen models achieve higher absolute performance, but retain the same preference for examples. Complete results by model, domain, and difficulty are in Appendices~\ref{app:cross-backbone}, \ref{app:domain}, and~\ref{app:difficulty}.

We also assess robustness to the examples used during the benchmark construction by evaluating with fresh examples that neither task generation nor filtering has seen. Although replacing the examples lowers absolute performance, these main findings remain consistent (Appendix~\ref{app:fresh-examples}).

\subsection{RQ2: What limits the effectiveness of retrieved knowledge?}
\label{sec:rq2}

RQ1 supplied knowledge directly from each task's target API. We now replace this oracle context with the top-five chunks retrieved by BGE-small~\citep{bge} from a FAISS index~\citep{faiss} over the knowledge bundles of all candidate APIs in the task's domain, including APIs without tasks. Each index contains only the knowledge components specified by its condition. We evaluate the same $1{,}670$ tasks with Qwen2.5-Coder-7B, keeping decoding and scoring unchanged.

\begin{figure}[t]
    \centering
    \includegraphics[width=0.95\linewidth]{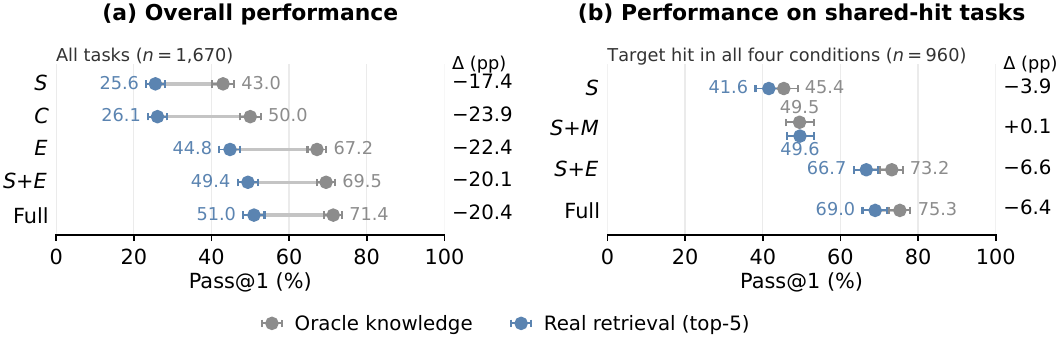}
    \caption{Oracle versus retrieved knowledge on Qwen2.5-Coder-7B. (a) Pass@1 on all $1{,}670$ tasks. (b) Pass@1 on the shared subset of $960$ tasks where the target API appears in the top-five retrieved chunks under all four conditions shown. Within each panel, oracle and retrieval results use the same tasks. $\Delta$ denotes retrieved minus oracle Pass@1 in percentage points, computed from unrounded values. Whiskers show $95\%$ API-cluster bootstrap intervals.}
    \label{fig:rq2-retrieval}
\end{figure}

\paragraph{Retrieval substantially reduces the gains from knowledge.}
Across the five conditions in Figure~\ref{fig:rq2-retrieval}a, retrieval lowers Pass@1 by $17.4$--$23.9$ percentage points. Full drops from $71.4\%$ with oracle knowledge to $51.0\%$ with retrieval. These losses exceed most differences between knowledge conditions in RQ1, showing that knowledge delivery strongly affects the gains achievable from the available content.

\paragraph{Successfully retrieving the target API knowledge does not close the gap.}
We next ask whether the performance drop is simply due to retrieval failing to find the target API. Figure~\ref{fig:rq2-retrieval}b considers the $960$ tasks where retrieval finds the target under all four conditions shown. Even on these tasks, retrieval lowers Pass@1 by $6.4$ percentage points for Full, $6.6$ for $S+E$, and $3.9$ for $S$, while $S+M$ remains essentially unchanged. Missing the target API therefore explains only part of the loss. \textit{The remaining gap suggests that retrieved knowledge is less effective than the complete target bundle, whether because only part of the bundle is retrieved or because it arrives alongside potentially distracting content.}

\paragraph{Examples remain valuable under retrieval.}
Full and $S+E$ retain their lead, followed by $E$, while $S$ and $C$ remain substantially lower. In particular, $S+E$ outperforms $S$ by $23.8$ percentage points. Examples therefore remain the most valuable knowledge component in this retrieval setting. 
Appendix~\ref{app:rq2-additional} reports all conditions and the detailed retrieval analysis.

\subsection{RQ3: What does parametric adaptation contribute?}
\label{sec:rq3}

We next examine whether parameter updates help models solve tasks involving novel APIs. We split the benchmark $80/20$ by target API and evaluate on $348$ held-out tasks covering $178$ APIs. We compare SFT, RAFT~\citep{zhang2024raft}, and three knowledge-editing methods, GRACE~\citep{grace}, MEMIT~\citep{memit}, and AlphaEdit-LoRA, a LoRA-based variant of AlphaEdit~\citep{alphaedit} (Appendix~\ref{app:setup-rq3}), against the untuned base model, both without external knowledge and with the target API's Full bundle.

\begin{table}[t]
\centering
\small
\caption{Pass@1 (\%) on $348$ held-out tasks.
$\Delta$ is the change from Base with the Full bundle
in percentage points, with $95\%$ API-cluster bootstrap
confidence intervals.
Full results are in Appendix~\ref{app:setup-rq3}.}
\label{tab:rq3-headline}
\begin{tabular}{lrrc}
\toprule
Method & Without Bundle & Full Bundle & $\Delta$ (95\% CI) \\
\midrule
Base           & 0.3  & 71.3          & --- \\
\midrule
SFT            & 2.3  & 82.8          & $+11.5\;[+6.8, +16.4]$ \\
RAFT           & 1.7  & \textbf{85.3} & $+14.1\;[+8.9, +19.2]$ \\
\midrule
GRACE          & 0.3  & 70.1          & $-1.1\;[-2.9, +0.3]$ \\
MEMIT          & 1.7  & 72.1          & $+0.9\;[-2.6, +4.4]$ \\
AlphaEdit-LoRA & 1.1  & 75.6          & $+4.3\;[-0.6, +9.4]$ \\
\bottomrule
\end{tabular}
\end{table}

\paragraph{Fine-tuning improves knowledge use on novel APIs.} With the Full knowledge bundle, SFT and RAFT raise Pass@1 from $71.3\%$ to $82.8\%$ and $85.3\%$, respectively (Table~\ref{tab:rq3-headline}). Without the bundle, both remain at or below $2.3\%$. Their gains on novel APIs therefore depend strongly on access to external knowledge: fine-tuning helps the model solve tasks with that knowledge, \textit{but does little to remove the need for it}.

\paragraph{The tested editing methods do not significantly improve knowledge-assisted performance on held-out APIs.} None of the editing methods significantly improves on the base model when the bundle is supplied. GRACE and MEMIT stay within about one point of the base model, and AlphaEdit-LoRA is $4.3$ points higher with an interval that includes zero. These results suggest that these editing methods are less effective than fine-tuning at enhancing the use of external knowledge.

\begin{figure}[t]
    \centering
    \includegraphics[width=\linewidth]{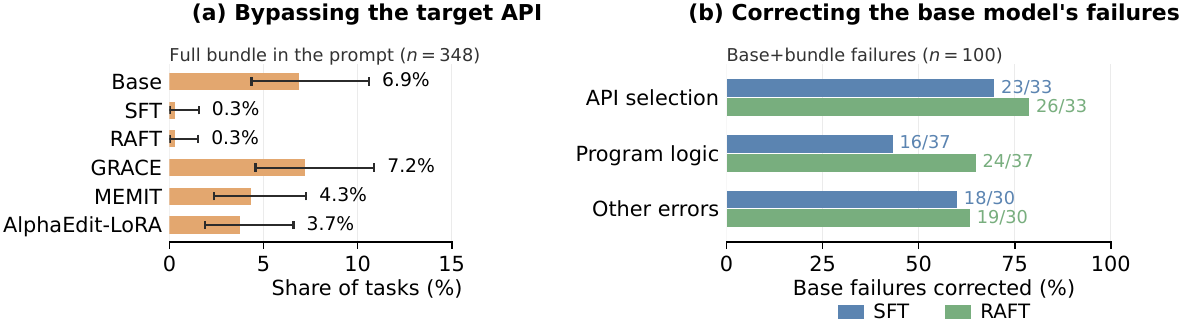}
    \caption{How adaptation changes model behavior with the Full knowledge bundle. (a) Percentage of the $348$ held-out tasks where models implement the functionality themselves instead of calling the target API, with $95\%$ API-cluster bootstrap confidence intervals. (b) Percentage of the base model's $100$ failures corrected by SFT and RAFT within each error category. Labels show corrected/total counts. Other errors comprise import, syntax, parameter, and shape/dtype errors.}
    \label{fig:rq3-bypass-fixes}
\end{figure}

\paragraph{Fine-tuning reduces selection and integration errors.}
With the Full bundle, SFT and RAFT reduce implementations that
bypass the target API from $6.9\%$ to $0.3\%$
(Figure~\ref{fig:rq3-bypass-fixes}a).
Of the base model's failures, they correct $23$ and $26$
of $33$ API-selection errors, respectively, and $16$ and $24$
of $37$ program-logic errors
(Figure~\ref{fig:rq3-bypass-fixes}b).
The gains therefore extend beyond choosing the API to
using it within the surrounding code.
Appendices~\ref{app:rq3-failures} and~\ref{app:rq3-examples}
provide error breakdowns and examples.

%% file: sec/discussions.tex
\paragraph{Organizing API knowledge for models.}
Our results suggest that API documentation should prioritize executable examples and explicit interface information. For unseen APIs, examples provide the strongest individual source of guidance. For updated APIs, signatures and parameter descriptions become especially important because an example may not reveal what has changed. Documentation for evolving libraries should therefore pair runnable examples with clear interface specifications and explicitly identify interface changes. Since $S+E$ approaches the performance of the Full bundle, concise, well-chosen knowledge can provide much of the benefit of more extensive documentation.

\paragraph{Retrieval quality extends beyond finding the target API.}
The gap between oracle and retrieved knowledge shows that finding the target API is only part of effective knowledge delivery. Even when retrieval succeeds, models can perform worse than when the target knowledge is supplied directly. Retrieval-augmented code evaluations should therefore measure both target-API recall and task success when the target is retrieved. For system design, this motivates attention to the completeness and clarity of relevant evidence, as well as interference from other retrieved content. Filtering distracting examples and training models to select relevant evidence are promising directions for making retrieved knowledge more useful.

\paragraph{Learning to use knowledge as APIs evolve.}
Our findings suggest a division of labor between external knowledge and parametric adaptation. \textit{External knowledge can keep models informed about changing API details, while parameter updates can improve how models select and apply that information.} Fine-tuning transfers to unseen APIs primarily when their knowledge is supplied, helping models call the target API and integrate the call into the surrounding code. This points to a training priority: strengthen the ability to reason over external API information and turn it into executable code. Frequently changing details, such as signatures and import paths, can then be supplied at inference time. The resulting approach combines up-to-date knowledge with reusable skills for applying it across APIs.

Our limitations are discussed in Appendix~\ref{app:limitations}.                                   

%% file: sec/conclusion.tex
We introduced \system{}, an automated diagnostic framework for studying how coding models learn to use novel APIs. By constructing benchmarks around model-specific knowledge gaps and linking controlled knowledge interventions to execution failures, it reveals what information models need and where its use breaks down. Our results show that usage examples provide the strongest standalone guidance, while explicit interface information is especially valuable for modified APIs. Retrieval leaves a performance gap even when the target API is found, and fine-tuning improves the use of supplied knowledge on unseen APIs without removing the need for that knowledge. As models and libraries evolve, \system{} provides a repeatable way to diagnose these complementary needs and guide adaptation.

%% file: app/a_limitations.tex
\section{Limitations}
\label{app:limitations}

\paragraph{Python libraries only.}
\label{app:lim-python}
Python's introspection facilities and executable library interfaces
support a common construction and validation pipeline across the
$21$ libraries and five domains studied here.
Extending \system{} to other languages requires language-specific
API extraction and execution support, and may require adapting the
knowledge schema and failure taxonomy.
Whether our findings transfer to different language
interfaces and documentation conventions remains untested.

\paragraph{Single target API per task.}
\label{app:lim-single-api}
Each task designates one target API while allowing helper APIs.
This design supports controlled comparisons of knowledge
interventions for a specific API.
It does not capture the additional challenges of selecting and
combining multiple unfamiliar APIs, which would require evaluating
interactions among knowledge bundles and diagnosing failures
across API boundaries.

\paragraph{Scope of the parametric study.}
\label{app:lim-parametric}
RQ3 compares LoRA-based SFT and RAFT with three knowledge-editing
methods on Qwen2.5-Coder-7B.
Using a common backbone and API split supports comparison within
this setting; the API-disjoint split specifically tests whether
adaptation transfers to held-out APIs.
Appendix~\ref{app:same-api} provides a complementary same-API
evaluation.
Our findings are limited to the tested backbone and configurations.
Other approaches, including full-parameter fine-tuning,
reinforcement learning, and continual pretraining, may yield
different trade-offs between internalizing API knowledge and
using external information.

\paragraph{API-level validation.}
\label{app:lim-harness}
Tasks span diverse program contexts without a common output interface, and stochastic behavior or environment-dependent numerical variation can further complicate direct output comparisons. Therefore, we validate target-API calls and require successful program execution, providing a consistent evaluation point aligned with our focus on API use (Appendix~\ref{app:bench-task}). These checks do not establish end-to-end correctness, since downstream outputs are not compared with the reference, and may reject valid alternative API usages. Broader validation would require task-level specifications that
capture both downstream behavior and equivalent API usages.

%% file: app/b_benchmark.tex
\section{Benchmark Construction Details}
\label{app:benchmark}

This appendix describes the construction pipeline, knowledge and task
schemas, and validation procedures. It also reports benchmark statistics,
release-boundary analyses, and examples.

\subsection{Knowledge Bundle Schema and Worked Example}
\label{app:bench-components}

\paragraph{Knowledge bundle.}
Table~\ref{tab:bundle-fields} summarizes the four knowledge components.
In model prompts, names and parameter descriptions appear as text,
examples and implementations as code blocks, and mechanism descriptions
as prose.

\begin{table}[htbp]
\centering
\small
\caption{Knowledge bundle fields and statistics over the $856$ target
APIs in the Qwen2.5-Coder-7B instance.}
\label{tab:bundle-fields}
\begin{tabularx}{\linewidth}{@{}l X X@{}}
\toprule
\textbf{Field} & \textbf{Content and source} & \textbf{Statistics} \\
\midrule
$S_{\text{name}}$
& Fully qualified API name, extracted from $v_{\text{new}}$.
& --- \\

$S_{\text{param}}$
& Parameter names, types, and defaults checked against the installed
signature, with descriptions from documentation.
& Median: $3$ parameters. \\

$E$
& Up to three LLM-generated usage examples grounded in API documentation,
with execution status checked in the sandbox
(\S\ref{app:bench-pipeline}).
& Mean: $2.5$ examples per API; $88.4\%$ execute successfully. \\

$M$
& Explanation of API behavior and design, grounded in a paper,
source code, or docstring.
& Median: $292$ words. Grounding source: paper ($139$ APIs),
source code ($376$), or docstring ($341$). \\

$C$
& Implementation and directly called same-module helpers,
with docstrings removed.
& Median: $26$ lines; maximum: $12{,}000$ characters. \\
\bottomrule
\end{tabularx}
\end{table}

\paragraph{Task schema.}
Each task is a JSON record with the fields in
Listing~\ref{lst:task-schema}. The model receives the description,
context code, and knowledge selected by the experimental condition.
The reference completion and test harness are hidden.

\begingroup
\lstset{
  basicstyle=\ttfamily\footnotesize,
  breaklines=true,
  breakatwhitespace=false,
  columns=fullflexible,
  keepspaces=true,
  showstringspaces=false
}

\begin{lstlisting}[
  caption={Task record fields.},
  label={lst:task-schema}
]
task_id             unique identifier
api_name            fully qualified target API name
library, domain     source library and domain
difficulty          "easy" | "medium" | "hard"
description         task specification, excluding the target API name
context_code        code preceding the masked region
masked_region       reference completion of the masked region
reference_solution  context code with the reference completion
test_harness        setup_code: target-call monitor
                    execution_test: reference-derived assertions
\end{lstlisting}

\paragraph{Worked example.}
Listing~\ref{lst:bundle-example} shows a bundle for
\texttt{kaiser\_sinc\_filter1d}, a newly introduced API in
\texttt{diffusers} that constructs a Kaiser-windowed sinc low-pass
filter. The fields provide complementary information:
$S_{\text{param}}$ describes the inputs;
$E$ shows how to construct and apply the filter;
$M$ explains its design; and $C$ exposes the computation.
Listing~\ref{lst:task-example} shows an easy task built from this bundle.

\begin{lstlisting}[
  caption={Abridged knowledge bundle for the filter API. Ellipses indicate omitted content.},
  label={lst:bundle-example}
]
S_name: diffusers.pipelines.ltx2.vocoder.kaiser_sinc_filter1d
S_param:
  - cutoff (float): Normalized frequency cutoff (relative to the sampling
    rate). Must be between 0 and 0.5 (the Nyquist frequency).
  - half_width (float): Used to determine the Kaiser window's beta
    parameter (affects transition width).
  - kernel_size (int): Size of the Kaiser window (and ultimately the
    Kaiser sinc kernel). Determines the length of the returned 1D filter.

E (3 examples, all executed; second shown):
  kernel = kaiser_sinc_filter1d(
      cutoff=0.25, half_width=0.02, kernel_size=101)
  weight = kernel.view(1, 1, -1)   # (out_channels, in_channels, k)
  pad = (weight.shape[-1] // 2, weight.shape[-1] // 2)
  filtered = F.conv1d(F.pad(signal, pad), weight)

M (source-grounded, 330 words):
  Implements a windowed sinc low-pass FIR kernel using a Kaiser window
  whose shape (beta) is chosen from an attenuation estimate derived from
  the desired transition width and kernel length. [...] half_width
  controls the transition band; the code multiplies it by 4 to form
  delta_f used in the Kaiser design formula [...] the windowed sinc is
  then normalized so the kernel sums to one (unity DC gain). [...]

C (21 lines):
  def kaiser_sinc_filter1d(cutoff: float, half_width: float,
                           kernel_size: int) -> torch.Tensor:
      delta_f = 4 * half_width
      half_size = kernel_size // 2
      amplitude = 2.285 * (half_size - 1) * math.pi * delta_f + 7.95
      if amplitude > 50.0:
          beta = 0.1102 * (amplitude - 8.7)
      [...]
      window = torch.kaiser_window(
          kernel_size, beta=beta, periodic=False)
      [...]
      filter = filter / filter.sum()
      return filter
\end{lstlisting}

\begin{lstlisting}[
  caption={Easy task for the filter API. The harness appears in Listing~\ref{lst:harness-example}.},
  label={lst:task-example}
]
description:
  Generate a compact, unity-gain low-pass FIR kernel suitable for audio
  anti-aliasing with a normalized cutoff of 0.25, a modest transition
  half-width of 0.05, and a kernel length of 101 samples. Return the
  kernel tensor and print its type, shape, dtype, and the first eight
  coefficients.

context_code:
  import torch
  cutoff = 0.25       # normalized frequency (0 < cutoff < 0.5)
  half_width = 0.05
  kernel_size = 101

masked_region:
  kernel = diffusers.pipelines.ltx2.vocoder.kaiser_sinc_filter1d(
      cutoff=cutoff, half_width=half_width, kernel_size=kernel_size)
  print(type(kernel))
  print(kernel.shape)
  print(kernel.dtype)
  print(kernel[:8])
\end{lstlisting}

\endgroup

\subsection{Pipeline Implementation}
\label{app:bench-pipeline}

Unless otherwise specified, construction uses GPT-5-mini with low
reasoning effort. Code executes in a network-isolated sandbox with
a $30$-second timeout and $16$\,GB memory limit, in a dedicated
environment containing the target library version $v_{\text{new}}$.
Table~\ref{tab:bench-funnel} summarizes the construction pipeline.

\begin{table}[htbp]
\centering
\small
\caption{APIs and tasks retained during construction.
Stages~1--3 and C1 are shared across backbones; C2 selects
backbone-specific task sets.
C3 checks scenario assertions, then the full harness.
The full-harness counts also include exclusions for unstable
reference records.
Execution-related exclusions affecting the 32B input and the
evaluated sets are detailed in \S\ref{app:bench-stats}.}
\label{tab:bench-funnel}
\begin{adjustbox}{max width=\linewidth}
\begin{tabular}{@{}lrrrrr@{}}
\toprule
\multicolumn{6}{l}{\textit{Shared stages}} \\
\midrule
Stage~1: candidate APIs
& \multicolumn{5}{l}{$2{,}600$ APIs from $21$ libraries} \\
Stage~2: knowledge bundles
& \multicolumn{5}{l}{$1{,}893$ APIs} \\
Stage~3: generated tasks
& \multicolumn{5}{l}{$2{,}330$ tasks for $1{,}014$ APIs from $20$ libraries} \\
C1: reference assertions
& \multicolumn{5}{l}{$2{,}329$ tasks} \\
\midrule
& & & \multicolumn{2}{c}{After C3} & \\
\cmidrule(lr){4-5}
Backbone
& C2 input
& After C2
& Assertions
& Full harness
& Evaluated \\
\midrule
Qwen2.5-Coder-7B
& $2{,}329$ & $2{,}145$ & $1{,}899$ & $1{,}671$ & $1{,}670$ \\
Qwen2.5-Coder-14B
& $2{,}329$ & $2{,}103$ & $1{,}867$ & $1{,}629$ & $1{,}628$ \\
Qwen2.5-Coder-32B
& $2{,}258$ & $2{,}028$ & $1{,}786$ & $1{,}569$ & $1{,}569$ \\
Seed-Coder-8B
& $2{,}329$ & $2{,}137$ & $1{,}905$ & $1{,}662$ & $1{,}661$ \\
OpenCoder-8B
& $2{,}329$ & $2{,}176$ & $1{,}950$ & $1{,}692$ & $1{,}691$ \\
DeepSeek-R1-Distill-Qwen-7B
& $2{,}329$ & $2{,}232$ & $1{,}986$ & $1{,}741$ & $1{,}740$ \\
\bottomrule
\end{tabular}
\end{adjustbox}
\end{table}

\paragraph{Stage 1: API discovery.}
We use 2023-12-01 as the primary release boundary for a shared
candidate pool. For each library, $v_{\text{old}}$ is the latest
PyPI release on or before this date, and $v_{\text{new}}$ is the
latest release at construction time
(Table~\ref{tab:bench-libraries}).
For \texttt{fastmcp} and \texttt{langgraph}, which have no release
before the boundary, we consider all public APIs.

We enumerate classes, functions, and methods in importable
submodules and record their signatures, docstrings, and source
locations. APIs are matched across versions by short name,
source file, and kind to accommodate import-path changes.
Unmatched APIs are labeled \emph{newly introduced}.
Matched APIs are labeled \emph{signature-modified} when parameters
are added or removed, or their names, kinds, or defaults change.
Changes to type annotations alone are ignored.

We exclude private or deprecated APIs, type aliases, externally
defined or vendored APIs, APIs without Python source, and those
with docstrings shorter than $40$ characters.
We also exclude cross-cutting private utilities and modules named
\texttt{base}, \texttt{main}, \texttt{impl}, or \texttt{internals}.
To limit repetitive candidates, we retain at most three classes
from groups of at least five sharing a CamelCase suffix
(e.g., \texttt{*Processor}).

We also design a candidate cap for some of the large libraries 
such as \texttt{diffusers} considering computational budget. 
Libraries exceeding their candidate cap ($100$--$300$) are
subsampled using maximal marginal relevance (MMR)~\citep{mmr} with MiniLM
embeddings~\citep{minilm} of API names and docstrings
($\lambda=0.5$).

The pool contains $2{,}584$ APIs that differ from the primary
reference release and $16$ additional APIs identified using the
later reference releases in \S\ref{app:release-boundary},
for a total of $2{,}600$ candidates.

\paragraph{Stage 2: Knowledge extraction.}
GPT-5-mini uses web search to extract surface information and
usage context from official documentation.
Parameter names, types, and defaults are checked against the
installed signature.
A separate call generates three usage examples from the signature,
parameter descriptions, and docstring.
Examples are labeled \emph{executed} if they run successfully,
\emph{static} if they pass syntax, target-reference, and import
checks but require unavailable resources, and \emph{failed}
otherwise.
If all three fail, generation is retried up to twice with
failure feedback. APIs without an executed or static example
are discarded.

The mechanism description $M$ is grounded in an associated paper
when available, or in the implementation for APIs with substantial
source code. Simple utilities and structural classes receive
a $50$--$100$-word description based on their docstrings.
The implementation component $C$ contains the API source and
directly called same-module helpers, with docstrings removed.
We exclude APIs primarily used for file I/O and those whose
implementation is unavailable or exceeds $12{,}000$ characters.
This stage yields $1{,}893$ complete bundles.

\paragraph{Stage 3: Task generation.}
For each bundle, one LLM call generates three tasks from the
API name, signature, parameter descriptions, examples, and
mechanism description.
The difficulty tiers vary the surrounding integration:
\emph{easy} tasks use a direct call with simple inputs;
\emph{medium} tasks place the API in a multi-step pipeline;
and \emph{hard} tasks build a more complex feature around it.
For signature-modified APIs, the prompt includes the old
signature and a change summary, and requires use of the
updated interface.
All tasks should describe realistic scenarios with plausible
inputs and distinguish the target from related APIs.

For APIs with only static examples, we instead generate three
easy variants: a minimal call, different inputs, and an edge case.
These account for $200$ tasks ($12.0\%$) across $92$ APIs
in the Qwen2.5-Coder-7B instance, including $192$ tasks from
\texttt{diffusers}.

We reject tasks that name the target API or an inflected variant
in the description, omit it from the reference code, or have
a reference solution that fails to execute.
Harnesses are constructed from reference executions
(\S\ref{app:bench-task}).
The resulting $2{,}330$ tasks cover $1{,}014$ APIs:
$1{,}087$ easy, $669$ medium, and $574$ hard.

\paragraph{Stage 4: Filtering.}
Tasks must satisfy three criteria:
\begin{itemize}
    \item \textbf{C1: Reference validity.}
    The reference solution must pass the scenario assertions
    and produce stable target-API call records
    (\S\ref{app:bench-task}).
    One task fails the scenario assertions, leaving 2,329 tasks in the C1 row of Table~\ref{tab:bench-funnel}. Three additional tasks have unstable reference records; their exclusion is accounted for in the Full harness column.

    \item \textbf{C2: Empirical novelty.}
    The target model receives only the context code and task
    description (Listing~\ref{lst:c2-prompt}).
    We sample three completions at $T=0.8$ and retain a task
    only if all three fail.
    This produces a separate task set for each backbone.

    \item \textbf{C3: Solvability with knowledge.}
    GPT-5-mini receives the task, the Full knowledge bundle
    ($S$, $E$, $M$, and $C$), and an explicit import line
    for the target API.
    Screening checks scenario assertions first, then the full
    harness with call-record validation
    (\S\ref{app:bench-task}).
    Each pass samples three completions at $T=0.2$ and requires
    at least one to pass.
\end{itemize}

Full-harness screening covers $2{,}124$ distinct tasks with stable
reference records, excluding the server task described in
\S\ref{app:bench-stats}.
It removes $299$ tasks, leaving $1{,}825$ distinct tasks
across the six evaluated instances.

\begin{lstlisting}[
  caption={C2 prompt without supplied API knowledge.},
  label={lst:c2-prompt},
  basicstyle=\ttfamily\footnotesize,
  breaklines=true,
  columns=fullflexible,
  keepspaces=true,
  showstringspaces=false
]
{context_code}

# Task:
{description}

# Complete the following code (output only the missing lines, no explanation):
\end{lstlisting}

\subsection{Task Format and Test Harness}
\label{app:bench-task}

\paragraph{Harness construction.}
An LLM generates six test scenarios that exercise the reference
solution, typically with inputs different from those in the task.
Each scenario reports output properties such as type, shape,
dtype, length, and scalar values.
We execute these scenarios and convert the recorded properties
into assertions, requiring at least two scenarios to execute
successfully.
Assertions that still pass when the target API is replaced by
a stub returning zeros or \texttt{None} are discarded.
These scenarios check library behavior on separate inputs;
the completion is evaluated through its own execution and
target-API call records.

\paragraph{Target-call monitor.}
Before execution, the harness wraps the target API with a
recording monitor, including references exposed through re-exported
module paths.
Classes are wrapped by subclassing to preserve
\texttt{isinstance} checks.
The call record is cleared after the context code runs, so only
calls from the completion are evaluated.
The harness requires the monitor to be installed successfully
and the completion to invoke the target API.
Implementations that reproduce the functionality without using
the target API are classified as \emph{bypass} in RQ3.

\paragraph{Call-record comparison.}
We execute the reference solution with four random seeds and
retain the fields of its target-API call records that remain
stable across runs.
Each reference call must have a matching call in the completion's
record.
We compare return-value fingerprints consisting of type, shape,
dtype, and a numerical checksum or content hash.
When return values are uninformative, such as for constructors,
decorators, or functions returning \texttt{None}, we compare
explicitly passed arguments instead.
Numeric types are normalized, omitted arguments take their
default values, and free-text strings are compared by type.

Among the $2{,}124$ tasks entering full-harness screening
(Table~\ref{tab:bench-funnel}), $44.8\%$ use return-value
comparisons and $42.1\%$ use argument comparisons.
For the remaining $13.1\%$, neither provides an informative
comparison, so the check verifies only a call or subclassing.
These checks assess target-API use within the program context;
they do not compare downstream outputs or establish end-to-end
program correctness (\S\ref{app:lim-harness}).

\paragraph{Evaluation.}
The completion replaces the masked region.
Execution proceeds through the target-module import, monitor
setup, context code, completion, call-record comparison, and
scenario assertions.
The harness also binds the target's bare name to the monitored
API when that name is undefined, supporting both bare-name and
fully qualified calls.
A task passes only if execution completes without error and
all checks succeed.
Listing~\ref{lst:harness-example} illustrates the harness for
the task in Listing~\ref{lst:task-example}.

\begin{lstlisting}[
  caption={Harness for the task in Listing~\ref{lst:task-example}.
  Monitor setup is omitted and the call-record comparison is
  summarized in comments.},
  label={lst:harness-example}
]
# Verify target-API use by the completion.
assert _target_api_spy_installed, \
    "target-API monitor was not installed"
assert _target_api_call_count[0] > 0, \
    "completion did not call the target API"

# Compare recorded calls with the reference fingerprint:
# torch.Tensor, shape (101,), dtype torch.float32,
# and the reference checksum.

# Check library behavior on a separate scenario.
from diffusers.pipelines.ltx2.vocoder import kaiser_sinc_filter1d

result = kaiser_sinc_filter1d(
    cutoff=0.20,
    half_width=0.04,
    kernel_size=201,
)
assert hasattr(result, 'shape') and \
    repr(result.shape) == 'torch.Size([201])'
assert result is not None and \
    (result.any() if hasattr(result, 'any') else len(result) > 0)
\end{lstlisting}

\subsection{Difficulty Tiers and Validation}
\label{app:difficulty-validation}

Stage~3 assigns difficulty tiers according to
\S\ref{app:bench-pipeline}.
We assess these labels through within-API comparisons of code
structure and blinded human rankings.

\paragraph{Code structure.}
We analyze $220$ APIs in the Qwen2.5-Coder-7B instance with
a retained task at each tier.
Table~\ref{tab:difficulty-structural} reports median code
properties and the proportion of triplets with a strict
easy-to-medium-to-hard increase.
The masked region grows from a median of $3$ to $7$ to $15$
lines, with a strict increase in $76.4\%$ of triplets.
AST size, function calls, and control-flow statements show
similar trends.
Harness assertion counts remain relatively stable, indicating
that the tiers primarily differ in the code to be completed
rather than the number of assertions.

\begin{table}[h]
\centering
\small
\caption{Code structure across $220$ within-API task triplets.
Tier columns report medians; the last column reports the
percentage with a strict easy-to-medium-to-hard increase
and its $95\%$ Wilson interval.
AST-based metrics exclude one triplet whose masked region
could not be parsed.}
\label{tab:difficulty-structural}
\begin{adjustbox}{max width=\linewidth}
\begin{tabular}{lrrrc}
\toprule
Metric & Easy & Medium & Hard & Increasing (\%, 95\% CI) \\
\midrule
Masked lines of code
& $3$ & $7$ & $15$ & $76.4\;[70.3, 81.5]$ \\
AST nodes in masked region
& $41$ & $77$ & $139$ & $80.4\;[74.6, 85.1]$ \\
Function calls
& $3$ & $6$ & $9$ & $52.5\;[45.9, 59.0]$ \\
Control-flow statements
& $0$ & $1$ & $2$ & $37.9\;[31.7, 44.5]$ \\
Context lines of code
& $8$ & $12$ & $16$ & $57.7\;[51.1, 64.1]$ \\
\midrule
Harness assertions
& $12$ & $13$ & $13$ & $5.5\;[3.1, 9.3]$ \\
\bottomrule
\end{tabular}
\end{adjustbox}
\end{table}

\paragraph{Human ranking.}
Two annotators independently ranked $50$ triplets sampled
from Stage~3 outputs, stratified by domain.
They saw each task's description, context code, and reference
completion, with tier labels, library names, and task identifiers
hidden and task order randomized.
They ranked tasks without ties by the reasoning and composition
needed to write the completion, assuming familiarity with
the API documentation.

Pairwise agreement with the assigned tiers was $94.7\%$ and
$98.0\%$ for the two annotators.
They agreed with each other on the complete ordering of
$92\%$ of triplets (Table~\ref{tab:difficulty-human}).
Together with the structural results, these rankings support
the tiers as measures of implementation complexity.

\begin{table}[h]
\centering
\small
\caption{Human ranking agreement on $50$ triplets.
Cohen's $\kappa$ is computed over per-task ranks;
brackets give $95\%$ bootstrap confidence intervals.}
\label{tab:difficulty-human}
\begin{adjustbox}{max width=\linewidth}
\begin{tabular}{lccc}
\toprule
Metric & Annotator 1 vs.\ tiers & Annotator 2 vs.\ tiers
& Between annotators \\
\midrule
Pairwise agreement (\%)
& $94.7$ & $98.0$ & $95.3$ \\
Identical triplet ranking (\%)
& $88$ & $94$ & $92$ \\
Mean Kendall's $\tau$
& $0.89$ & $0.96$ & $0.91$ \\
Cohen's $\kappa$
& $0.86\;[0.75, 0.96]$
& $0.94\;[0.86, 1.00]$
& $0.89\;[0.78, 0.98]$ \\
\bottomrule
\end{tabular}
\end{adjustbox}
\end{table}

\subsection{Benchmark Statistics}
\label{app:bench-stats}

\paragraph{Primary instance.}
Table~\ref{tab:bench-libraries} breaks down the Qwen2.5-Coder-7B instance by library. It contains $1{,}670$ tasks for $856$ APIs from $20$ libraries. Of these APIs, $688$ are newly introduced and $168$ are signature-modified, covering $1{,}376$ and $294$ tasks, respectively. Signature-modified APIs are concentrated in libraries with long-lived public interfaces, most notably \texttt{scipy} ($66$ of its $89$ APIs).

\begin{table}[h]
\centering\small
\caption{The Qwen2.5-Coder-7B instance by library. $v_{\text{old}}$ is the latest release on or before the 2023-12-01 boundary (``---'': no such release) and $v_{\text{new}}$ is the release used for construction. \emph{Mod.}\ counts signature-modified APIs; E/M/H count easy, medium, and hard tasks.}
\label{tab:bench-libraries}
\setlength{\tabcolsep}{4pt}
\begin{adjustbox}{max width=\linewidth}
\begin{tabular}{llllrrrc}
\toprule
\textbf{Domain} & \textbf{Library} & $v_{\text{old}}$ & $v_{\text{new}}$ & \textbf{APIs} & \textbf{Mod.} & \textbf{Tasks} & \textbf{E / M / H} \\
\midrule
\multirow{2}{*}{Agent tools}
 & \texttt{fastmcp}     & ---       & 3.2.4     & $111$ & $0$ & $219$ & $93$ / $68$ / $58$ \\
 & \texttt{langgraph}   & ---       & 1.1.10    & $25$ & $0$ & $49$ & $23$ / $17$ / $9$ \\
\midrule
\multirow{7}{*}{AI for science}
 & \texttt{ase}         & 3.22.1    & 3.28.0    & $73$ & $8$ & $128$ & $56$ / $36$ / $36$ \\
 & \texttt{astropy}     & 6.0.0     & 7.2.0     & $56$ & $15$ & $101$ & $39$ / $40$ / $22$ \\
 & \texttt{deepchem}    & 2.7.1     & 2.8.0     & $60$ & $0$ & $110$ & $46$ / $34$ / $30$ \\
 & \texttt{MDAnalysis}  & 2.6.1     & 2.10.0    & $11$ & $2$ & $23$ & $11$ / $6$ / $6$ \\
 & \texttt{pymatgen}    & 2023.11.12 & 2026.3.23 & $4$ & $2$ & $6$ & $3$ / $3$ / $0$ \\
 & \texttt{rdkit}       & 2023.9.2  & 2026.3.1  & $6$ & $0$ & $15$ & $6$ / $6$ / $3$ \\
 & \texttt{scanpy}      & 1.9.6     & 1.11.5    & $5$ & $3$ & $7$ & $3$ / $2$ / $2$ \\
\midrule
\multirow{3}{*}{Data science}
 & \texttt{numpy}       & 1.26.2    & 2.4.4     & $19$ & $13$ & $27$ & $10$ / $8$ / $9$ \\
 & \texttt{pandas}      & 2.1.3     & 3.0.2     & $26$ & $11$ & $45$ & $18$ / $18$ / $9$ \\
 & \texttt{scipy}       & 1.11.4    & 1.17.1    & $89$ & $66$ & $166$ & $67$ / $59$ / $40$ \\
\midrule
\multirow{3}{*}{Deep learning}
 & \texttt{diffusers}   & 0.24.0    & 0.38.0    & $161$ & $13$ & $335$ & $258$ / $43$ / $34$ \\
 & \texttt{torch}       & 2.1.1     & 2.11.0    & $27$ & $6$ & $54$ & $23$ / $16$ / $15$ \\
 & \texttt{transformers} & 4.35.2   & 5.7.0     & $147$ & $22$ & $311$ & $136$ / $101$ / $74$ \\
\midrule
\multirow{5}{*}{Software eng.}
 & \texttt{django}      & 4.2.7     & 5.2.13    & $17$ & $2$ & $33$ & $13$ / $9$ / $11$ \\
 & \texttt{fastapi}     & 0.104.1   & 0.136.1   & $4$ & $1$ & $8$ & $3$ / $3$ / $2$ \\
 & \texttt{flask}       & 3.0.0     & 3.1.3     & $1$ & $1$ & $3$ & $1$ / $1$ / $1$ \\
 & \texttt{pydantic}    & 2.5.2     & 2.13.3    & $5$ & $2$ & $10$ & $4$ / $4$ / $2$ \\
 & \texttt{sqlalchemy}  & 2.0.23    & 2.0.49    & $9$ & $1$ & $20$ & $9$ / $7$ / $4$ \\
\midrule
\multicolumn{4}{l}{\textbf{Total}} & $\mathbf{856}$ & $\mathbf{168}$ & $\mathbf{1{,}670}$ & $822$ / $481$ / $367$ \\
\bottomrule
\end{tabular}
\end{adjustbox}
\end{table}

\paragraph{Instances for all backbones.}
Table~\ref{tab:bench-backbones} summarizes the six
model-conditioned instances.
Model-specific C2 filtering yields $1{,}825$ distinct tasks,
of which $1{,}405$ are shared by all six instances and used
for cross-backbone comparisons. 
Two execution constraints affect coverage.
For Qwen2.5-Coder-32B, all $71$ candidate \texttt{torch} tasks
are excluded before C2 because distributed-execution workloads
exceed the cluster's memory limit.
We also specially exclude a task targeting
\texttt{torch.distributed.debug.stop\_debug\_server}
where present, because its server process persists beyond the
sandbox and prevents reliable scoring.
This accounts for the difference between the \emph{Full harness}
and \emph{Evaluated} columns in Table~\ref{tab:bench-funnel}.

\begin{table}[h]
\centering\small
\caption{Model-conditioned benchmark instances. \emph{Mod.}\ is the share of tasks whose target API is signature-modified.}
\label{tab:bench-backbones}
\begin{adjustbox}{max width=\linewidth}
\begin{tabular}{lrrrcr}
\toprule
\textbf{Backbone} & \textbf{Tasks} & \textbf{APIs} & \textbf{Libraries} & \textbf{E / M / H} & \textbf{Mod.} \\
\midrule
Qwen2.5-Coder-7B-Instruct   & $1{,}670$ & $856$ & $20$ & $822$ / $481$ / $367$ & $17.6\%$ \\
Qwen2.5-Coder-14B-Instruct  & $1{,}628$ & $839$ & $20$ & $805$ / $472$ / $351$ & $16.4\%$ \\
Qwen2.5-Coder-32B-Instruct  & $1{,}569$ & $804$ & $19$ & $779$ / $459$ / $331$ & $16.3\%$ \\
Seed-Coder-8B-Instruct      & $1{,}661$ & $850$ & $20$ & $817$ / $485$ / $359$ & $17.6\%$ \\
OpenCoder-8B-Instruct       & $1{,}691$ & $869$ & $20$ & $831$ / $489$ / $371$ & $19.0\%$ \\
DeepSeek-R1-Distill-Qwen-7B & $1{,}740$ & $886$ & $20$ & $859$ / $507$ / $374$ & $18.9\%$ \\
\midrule
Shared by all six           & $1{,}405$ & ---   & ---  & $702$ / $418$ / $285$ & --- \\
\bottomrule
\end{tabular}
\end{adjustbox}
\end{table}

\subsection{Sensitivity to Model-Specific Release Boundaries}
\label{app:release-boundary}
We use a shared release boundary to retain a broader set of
tasks involving recent APIs, including those released before
a backbone's nominal cutoff but still challenging for that
model under C2.
This captures potential gaps in API knowledge that release
dates alone may overlook, increasing benchmark coverage.

We test sensitivity to this choice by restricting each instance
to APIs introduced or signature-modified after a
backbone-specific boundary: the documented training-data cutoff
when available, or the model's release date otherwise
(Table~\ref{tab:release-share}).
We call these tasks \emph{release-novel} and recompute the analyses
with the same statistical procedures, relabeling novelty types
relative to each boundary.
For RQ3, training is unchanged; only the test set is restricted,
from $348$ to $241$ tasks.

\begin{table}[h]
\centering
\small
\caption{Tasks retained under backbone-specific release boundaries.
OpenCoder uses its documented training-data cutoff;
the other models use their release dates.
The restriction also changes library coverage.}
\label{tab:release-share}
\begin{tabular}{llrr}
\toprule
Backbone & Boundary & Retained tasks & Share (\%) \\
\midrule
OpenCoder-8B       & 2023-12-01 & $1{,}679$ & $99.3$ \\
Qwen2.5-Coder-7B   & 2024-09-01 & $1{,}166$ & $69.8$ \\
Qwen2.5-Coder-14B  & 2024-11-12 & $1{,}099$ & $67.5$ \\
Qwen2.5-Coder-32B  & 2024-11-12 & $1{,}075$ & $68.5$ \\
R1-Distill-Qwen-7B & 2025-01-20 & $1{,}025$ & $58.9$ \\
Seed-Coder-8B      & 2025-06-04 & $801$     & $48.2$ \\
\bottomrule
\end{tabular}
\end{table}

\paragraph{Main findings.}
The main patterns persist on the release-novel subsets
(Table~\ref{tab:release-headline}).
Examples outperform surface information on all six backbones,
and the oracle-to-retrieval gap remains similar.
With the Full bundle, SFT and RAFT retain substantial gains,
while none of the tested editing methods significantly
outperforms the base model.
Same-API training also retains its advantage over cross-API
training without external knowledge.

Of $79$ paired contrasts, $73$ retain their significance status.
The novelty-type interactions remain positive across all
backbones, though some lose significance on R1-Distill and
Seed-Coder.
Thus, the core findings persist, though statistical support
for some secondary comparisons depends on the task subset.

\begin{table}[h]
\centering
\small
\caption{Results on the full instances and release-novel subsets.
Unless indicated otherwise, results use Qwen2.5-Coder-7B.
Differences are in percentage points, with $95\%$
API-cluster bootstrap confidence intervals.
For the cross-backbone rows, counts indicate the number
of backbones with a significant effect.}
\label{tab:release-headline}
\setlength{\tabcolsep}{4pt}
\begin{adjustbox}{max width=\linewidth}
\begin{tabular}{llcc}
\toprule
& Comparison & Full instance & Release-novel \\
\midrule
\multirow{4}{*}{RQ1}
& $E-S$
& $+24.2\;[+21.2, +27.3]$
& $+23.8\;[+20.1, +27.6]$ \\
& $E>S$: significant backbones
& $6/6$ & $6/6$ \\
& $(E-S)_{\mathrm{new}}>(E-S)_{\mathrm{mod}}$:
  significant backbones
& $6/6$ & $5/6$ \\
& $(SE-E)_{\mathrm{mod}}>(SE-E)_{\mathrm{new}}$:
  significant backbones
& $6/6$ & $4/6$ \\
\midrule
RQ2
& Full: retrieved $-$ oracle
& $-20.4\;[-23.2, -17.7]$
& $-20.8\;[-24.1, -17.7]$ \\
\midrule
\multirow{4}{*}{RQ3}
& SFT $-$ Base, with Full
& $+11.5\;[+6.8, +16.4]$
& $+10.0\;[+4.6, +15.4]$ \\
& RAFT $-$ Base, with Full
& $+14.1\;[+8.9, +19.2]$
& $+11.2\;[+5.4, +17.2]$ \\
& MEMIT $-$ Base, with Full
& $+0.9\;[-2.6, +4.4]$
& $+1.2\;[-3.0, +5.8]$ \\
& AlphaEdit-LoRA $-$ Base, with Full
& $+4.3\;[-0.6, +9.4]$
& $+2.5\;[-3.6, +8.7]$ \\
\midrule
Same-API
& Same-API $-$ cross-API, without bundle
& $+17.6\;[+14.7, +20.7]$
& $+18.5\;[+15.1, +22.5]$ \\
\bottomrule
\end{tabular}
\end{adjustbox}
\end{table}

\subsection{Human Evaluation of Descriptions and Tasks}
\label{app:human-eval-quality}

Three annotators independently evaluated mechanism descriptions
and tasks generated by the Stage-2 and
Stage-3 generators.
Each reviewer had research experience with at least three
of the sampled libraries.
Items were rated on three dimensions using a five-point
Likert scale.

\paragraph{Mechanism descriptions.}
\label{app:human-eval-mdesc}
We sampled $75$ bundles, with $15$ per domain and balanced
coverage across libraries within each domain.
Reviewers assessed each mechanism description for
\emph{accuracy} of its claims about API behavior,
\emph{specificity} to the target API, and
\emph{groundedness} in the docstring, source code, or cited paper.
Mean scores range from $4.28$ to $4.46$, with all per-domain
means at least $4.0$ (Table~\ref{tab:human-quality}).

\paragraph{Tasks.}
\label{app:human-eval-task}
We sampled $50$ tasks, with $10$ per domain.
Reviewers assessed \emph{description clarity},
\emph{scenario realism}, and \emph{harness alignment}:
whether an implementation that meets the description and
correctly uses the target API would pass the checks.
Mean scores range from $4.56$ to $4.75$.

Ratings are concentrated at $4$ and $5$, indicating favorable
assessments overall.
However, low Krippendorff's $\alpha$ indicates limited
agreement beyond chance, so small differences between
mean scores should not be overinterpreted.

\begin{table}[h]
\centering
\small
\caption{Human ratings of $75$ mechanism descriptions and
$50$ tasks by three reviewers on a five-point scale.
Within $\pm1$ is the mean pairwise proportion of ratings
differing by at most one point.
$\alpha$ denotes interval Krippendorff's alpha.}
\label{tab:human-quality}
\begin{tabular}{llrrrr}
\toprule
Item & Dimension & Mean & SD & Within $\pm1$ & $\alpha$ \\
\midrule
\multirow{3}{*}{Mechanism $M$}
 & Accuracy            & $4.33$ & $0.65$ & $90.7\%$ & $-0.07$ \\
 & Specificity         & $4.46$ & $0.73$ & $90.7\%$ & $0.22$ \\
 & Groundedness        & $4.28$ & $0.81$ & $80.0\%$ & $0.01$ \\
\midrule
\multirow{3}{*}{Task}
 & Description clarity & $4.75$ & $0.69$ & $89.3\%$ & $-0.12$ \\
 & Scenario realism    & $4.56$ & $0.60$ & $97.3\%$ & $0.19$ \\
 & Harness alignment   & $4.66$ & $0.78$ & $83.3\%$ & $-0.14$ \\
\bottomrule
\end{tabular}
\end{table}

\subsection{Sample Tasks}
\label{app:bench-samples}

Listings~\ref{lst:sample-hard}--\ref{lst:sample-new} show three further tasks from the Qwen2.5-Coder-7B instance: the hard task generated from the bundle in Listing~\ref{lst:bundle-example}, an easy task for a signature-modified API, and a medium task for a newly introduced API from another domain.

\begin{lstlisting}[caption={Hard task for \texttt{kaiser\_sinc\_filter1d} (abridged).},label={lst:sample-hard}]
description:
  Implement a simple multirate downsampling pipeline for an audio
  dataset: given several long mono recordings (shape (N, T)), low-pass
  filter each signal to avoid aliasing and then decimate by an integer
  factor M=4. The anti-aliasing filter must have cutoff at 0.5/M [...], a
  half-width of 0.02, and kernel size of 161. Process each recording in
  contiguous, non-overlapping frames of 8192 samples to keep memory
  bounded, apply zero-phase centering by symmetric padding during
  convolution [...], and concatenate results to produce downsampled
  recordings. [...]
context_code:
  import torch
  import torch.nn.functional as F
  sr = 16000
  records = [...]   # three synthetic recordings of 8-12.5 seconds
  M = 4
  cutoff = 0.5 / M
  half_width = 0.02
  kernel_size = 161
  frame_size = 8192
masked_region:
  kernel = diffusers.pipelines.ltx2.vocoder.kaiser_sinc_filter1d(
      cutoff=cutoff, half_width=half_width, kernel_size=kernel_size)
  weight = kernel.view(1, 1, -1)
  pad = (weight.shape[-1] // 2, weight.shape[-1] // 2)
  downsampled = []
  for rec in records:
      out_frames = []
      for f in rec.split(frame_size):
          f = f.unsqueeze(0).unsqueeze(0)
          filtered = F.conv1d(F.pad(f, pad), weight).squeeze(0).squeeze(0)
          out_frames.append(filtered[::M])
      downsampled.append(torch.cat(out_frames, dim=0))
  [...]
\end{lstlisting}

\begin{lstlisting}[caption={Easy task for the signature-modified API \texttt{diffusers.models.embeddings.get\_2d\_sincos\_pos\_embed\_from\_grid}. Relative to the release before the boundary, whose signature was \texttt{(embed\_dim, grid)}, the function gained an \texttt{output\_type} argument that defaults to NumPy output.},label={lst:sample-modified}]
description:
  In a training data preprocessing step for a Vision Transformer, generate
  a 2-D sinusoidal positional embedding for a 7x7 patch grid and return it
  as a PyTorch tensor on the CPU so it can be concatenated directly with
  torch-based patch features. The embedding must have dimensionality 64
  and be returned as torch.Tensor (not numpy). Code written for the
  previous library version that only accepted (embed_dim, grid) will fail
  here unless the new output-type parameter is provided.
context_code:
  import torch
  import numpy as np
  H, W = 7, 7
  ys, xs = torch.meshgrid(torch.arange(H), torch.arange(W), indexing='ij')
  grid = torch.stack([xs, ys], dim=0).reshape(2, -1)   # shape (2, H*W)
  embed_dim = 64
masked_region:
  pos_embed = diffusers.models.embeddings.get_2d_sincos_pos_embed_from_grid(
      embed_dim=embed_dim, grid=grid, output_type='pt')
  print(type(pos_embed), pos_embed.shape)
\end{lstlisting}

\begin{lstlisting}[caption={Medium task for the newly introduced API \texttt{pandas.col} (\texttt{pandas} 3.0).},label={lst:sample-new}]
description:
  Given a ride-sharing trip log with trip durations (seconds) and rider
  IDs, compute each rider's average trip duration, then add a boolean
  indicator 'heavy_user' to the original DataFrame: True for rows whose
  rider's average duration exceeds 1800 seconds (30 minutes). Use a
  deferred expression to perform the boolean selection when marking rows
  in place.
context_code:
  import pandas as pd
  trips = pd.DataFrame({
      "rider_id": [1, 2, 1, 3, 2, 1],
      "trip_id": [11, 12, 13, 14, 15, 16],
      "duration_s": [1200, 2400, 3600, 600, 3000, 1800]})
  avg_per_rider = (trips.groupby("rider_id")["duration_s"].mean()
                   .rename("avg_duration_s").reset_index())
  trips = trips.merge(avg_per_rider, on="rider_id")
  trips["heavy_user"] = False
masked_region:
  trips.loc[pandas.col("avg_duration_s") > 1800, "heavy_user"] = True
\end{lstlisting}

%% file: app/c_failtax.tex
\section{Failure Classification}
\label{app:failtax}

\subsection{Definitions and Labeling}
\label{app:failtax-rules}

Each failed greedy completion receives one of six labels
(Section~\ref{sec:failtax}):
\begin{itemize}
    \item \texttt{WrongAPISelection}: the completion uses a different
    API, reimplements the functionality, or makes no target-API call.
    This category also includes outputs with no extractable code.

    \item \texttt{WrongImport}: the completion refers to the target
    API but cannot resolve its name, module path, or import.

    \item \texttt{WrongSyntax}: the completion raises
    \texttt{SyntaxError} or \texttt{IndentationError}.

    \item \texttt{WrongParam}: the target API rejects the supplied
    arguments at its interface, such as an unexpected keyword,
    a missing argument, incorrect arity, or an invalid primitive type.

    \item \texttt{WrongShapeDtype}: the target API accepts the
    argument signature but rejects the input structure, such as
    tensor shape, array dtype, data-frame columns, or schema.

    \item \texttt{WrongLogic}: the failure concerns the surrounding
    computation, such as incorrectly prepared inputs, misuse of
    a return value, an invalid callback, or incorrect control flow.
    This includes calls that fail to match the reference record
    despite otherwise acceptable arguments.
\end{itemize}

The first three labels form the \emph{API-level} group.
Classification follows the order API-level, parameter,
shape/dtype, then logic, using the location and cause of the
failure to distinguish categories.
For example, an error after the target call returns is a logic
failure, not a parameter error.

\paragraph{Rule-based labeling.}
Syntax and indentation errors receive \texttt{WrongSyntax}.
Completions that finish execution without calling the target API
receive \texttt{WrongAPISelection}.
A \texttt{NameError} identifying the target's leaf name or
top-level package receives \texttt{WrongImport}.
Other failures are classified using an LLM.

\paragraph{LLM labeling.}
GPT-5-mini (temperature $0$, low reasoning effort) receives
the task description, target API, reference-call summary,
generated code, AST-extracted call information, exception type,
undefined name if present, and the final $500$ characters
of the traceback.
It selects one of the four groups and provides a brief rationale
(Listing~\ref{lst:classifier-prompt}).
API-level predictions are then mapped to the three corresponding
labels using the exception and rationale.
Classification applies only to failed completions and does not
affect Pass@1.

Timeouts are represented by a \texttt{Timeout} exception;
call-record mismatches use the check's assertion message.
For malformed classifier outputs, the fallback label is
\texttt{WrongLogic} if the code mentions the target's leaf name,
and \texttt{WrongAPISelection} otherwise.

\begin{lstlisting}[
  caption={Failure-classifier prompt, with category rules abridged.
  Full rules are provided in the released code.},
  label={lst:classifier-prompt},
  basicstyle=\ttfamily\footnotesize,
  breaklines=true,
  columns=fullflexible,
  keepspaces=true,
  showstringspaces=false
]
System: You classify a single failed code generation into exactly ONE
category. Output only compact JSON: {"class": <CLASS>, "rationale": <one
sentence>}.

Task description: {description}
Target API: {target_api}
Reference call (from reference solution): {reference_call}
Predicted code (pass@1 sample):
```python
{predicted_code}

AST-extracted: api_called_leaf={leaf}, kwargs={kwargs}
Exception type from execution: {exception_type}
Undefined symbol from stderr: {undefined_name}
Stderr (last 500 chars):
{stderr}

Pick EXACTLY ONE of: W_API, W_PARAM, W_SHAPE_DTYPE, W_LOGIC.
Apply these rules IN ORDER and pick the FIRST that matches.

W_API - the model got the API NAME wrong, failed to name it at all, or
failed to emit code that could invoke it: [...]
W_PARAM - right API; the TARGET API's OWN signature rejected the
arguments: [...]
W_SHAPE_DTYPE - right API, right kwarg names; failure is a NUMERIC /
STRUCTURAL array-like contract violation raised INSIDE the target API
[...]
W_LOGIC - right API, plausible args, surrounding program is wrong: [...]
Output only: {"class": <CLASS>, "rationale": <one sentence citing the specific signal>}
\end{lstlisting}

\subsection{Examples}
\label{app:failtax-examples}

Table~\ref{tab:failtax-examples} illustrates each category using
Qwen2.5-Coder-7B completions with the Full bundle.

\begin{table}[h]
\centering
\footnotesize
\caption{Failure examples with the Full bundle.
Completion excerpts show the relevant operation.}
\label{tab:failtax-examples}
\setlength{\tabcolsep}{4pt}
\begin{adjustbox}{max width=\linewidth}
\begin{tabular}{
  >{\raggedright\arraybackslash}p{3.5cm}
  >{\raggedright\arraybackslash}p{5.7cm}
  >{\raggedright\arraybackslash}p{5.1cm}}
\toprule
Label and target & Completion & Failure signal \\
\midrule
\texttt{WrongAPISelection}\newline
\texttt{AzureProvider} (\texttt{fastmcp})
&
Builds the identifier list manually without constructing
\texttt{AzureProvider}.
&
The monitor records no target-API call.
\\
\midrule
\texttt{WrongImport}\newline
\texttt{push\_notification} (\texttt{fastmcp})
&
\texttt{notifications.push\_notification(}\newline
\texttt{~~session\_id, notification, docket)}
&
\texttt{NameError}: the module alias
\texttt{notifications} is undefined.
\\
\midrule
\texttt{WrongSyntax}\newline
\texttt{is\_enabled} (\texttt{fastmcp})
&
\texttt{if visibility.is\_enabled(component):}\newline
\texttt{~~return component.label}
&
\texttt{SyntaxError}: \texttt{return} outside a function.
\\
\midrule
\texttt{WrongParam}\newline
\texttt{black\_tophat} (\texttt{scipy})
&
\texttt{out = ndimage.black\_tophat(traces,}\newline
\texttt{~~structure=structure, axis=1)}
&
\texttt{TypeError}: unexpected keyword argument
\texttt{axis}.
\\
\midrule
\texttt{WrongShapeDtype}\newline
\texttt{FastMCPComponent} (\texttt{fastmcp})
&
Calls \texttt{FastMCPComponent.model\_validate}
with \texttt{config}.
&
Pydantic \texttt{ValidationError}: the input schema
does not permit the field \texttt{owner}.
\\
\midrule
\texttt{WrongLogic}\newline
\texttt{validate\_parse\_dates\_presence}
(\texttt{pandas})
&
Uses the set returned by the target API to index
a data frame with \texttt{df[parsed]}.
&
The API call succeeds, but indexing raises
\texttt{TypeError}: a set is not a supported indexer.
\\
\bottomrule
\end{tabular}
\end{adjustbox}
\end{table}

\subsection{Human Evaluation}
\label{app:human-eval-failtax}

Two annotators independently labeled shared 61 failure samples 
sampled across all five domains and knowledge conditions, without seeing
the classifier's predictions. We report both six-category agreement and four-group agreement,
which merges the three API-level labels.

\paragraph{Agreement between annotators.}
On the $61$ shared items, agreement was $90.2\%$
(Cohen's $\kappa=0.88$) for the six categories and
$96.7\%$ ($\kappa=0.93$) for the four groups.

\paragraph{Agreement with the classifier.}
The classifier agrees with the annotators on $90$--$93\%$
of four-group labels and approximately $84\%$ of six-category
labels (Table~\ref{tab:failtax-vs-llm}).
Against Annotator B, \texttt{WrongAPISelection} has precision
$0.95$ and recall $1.00$.
The less reliable categories are \texttt{WrongImport}
(precision $0.69$, recall $0.90$) and
\texttt{WrongShapeDtype} (precision $0.65$, recall $0.79$),
most often confused with \texttt{WrongSyntax} and
\texttt{WrongLogic}, respectively.

\begin{table}[h]
\centering
\small
\caption{Agreement between the classifier and each annotator.
Six-category comparisons include only items assigned a specific
category by the annotator; four-group comparisons also include
items labeled only as API-level.}
\label{tab:failtax-vs-llm}
\begin{tabular}{llrrr}
\toprule
Annotator & Categories & $n$ & Agreement (\%) & $\kappa$ \\
\midrule
A & Four & $88$  & $93.2$ & $0.89$ \\
A & Six  & $81$  & $84.0$ & $0.81$ \\
B & Four & $143$ & $90.2$ & $0.82$ \\
B & Six  & $92$  & $83.7$ & $0.80$ \\
\bottomrule
\end{tabular}
\end{table}

%% file: app/d_setup.tex
\section{Detailed Experimental Setup}
\label{app:setup}

\subsection{Models and Decoding}
\label{app:setup-models}

We evaluate six instruction-tuned checkpoints using their native
chat templates (Table~\ref{tab:setup-backbones}).
Pass@1 uses greedy decoding ($T=0$), a repetition penalty of
$1.0$, and a context limit of $16{,}384$ tokens
($8{,}192$ for OpenCoder-8B).
The generation budget is $2{,}048$ tokens, except for
DeepSeek-R1-Distill-Qwen-7B, which receives $8{,}192$ tokens
to accommodate its reasoning trace.
Inference uses BF16 with vLLM; GRACE uses the HuggingFace
backend to support its runtime layer hook.
For Pass@5, we generate $20$ samples at $T=0.8$ and apply
the unbiased estimator of \citet{chen2021codex}.

\paragraph{Code extraction.}
All models and conditions use the same extraction procedure.
After removing reasoning traces, we select the longest
parseable Python-tagged code block, followed by the longest
parseable block with any tag.
If neither is available, we attempt to parse a single
unterminated block or an entirely unfenced response.
If extraction fails, we repeat the procedure on the text
following the last repeated completion instruction.
Responses with no extractable code are treated as empty
completions.
The extracted code replaces the masked region and is evaluated
using the harness in Appendix~\ref{app:bench-task}.

\begin{table}[h]
\centering
\small
\caption{Model checkpoints and maximum generation lengths.
Qwen2.5-Coder-7B is the primary backbone.}
\label{tab:setup-backbones}
\setlength{\tabcolsep}{4pt}
\begin{adjustbox}{max width=\linewidth}
\begin{tabular}{llr}
\toprule
Backbone & Checkpoint & New tokens \\
\midrule
Qwen2.5-Coder-7B
& \texttt{Qwen/Qwen2.5-Coder-7B-Instruct}
& $2{,}048$ \\
Qwen2.5-Coder-14B
& \texttt{Qwen/Qwen2.5-Coder-14B-Instruct}
& $2{,}048$ \\
Qwen2.5-Coder-32B
& \texttt{Qwen/Qwen2.5-Coder-32B-Instruct}
& $2{,}048$ \\
OpenCoder-8B
& \texttt{infly/OpenCoder-8B-Instruct}
& $2{,}048$ \\
Seed-Coder-8B
& \texttt{ByteDance-Seed/Seed-Coder-8B-Instruct}
& $2{,}048$ \\
DeepSeek-R1-Distill-Qwen-7B
& \texttt{deepseek-ai/DeepSeek-R1-Distill-Qwen-7B}
& $8{,}192$ \\
\bottomrule
\end{tabular}
\end{adjustbox}
\end{table}

\subsection{Knowledge Conditions and Prompts}
\label{app:knowledge-conditions}

We vary the supplied knowledge across twelve conditions:
no knowledge, $S_{\text{name}}$, $S$, $E$, $M$, $C$,
$S{+}M$, $S{+}E$, $S{+}C$, $S{+}M{+}C$,
$S{+}E{+}C$, and Full.
Here, $S$ includes the API name and parameter information
(types, defaults, and descriptions); $S_{\text{name}}$
provides only the name.
The $36$ pairwise comparisons in
Appendix~\ref{app:full-conditions} cover nine conditions:
no knowledge, $S$, $E$, $M$, $C$, $S{+}M$, $S{+}E$,
$S{+}C$, and Full.

Selected components appear under a
\texttt{Reference Documentation} header in a fixed order:
name, parameters, examples, mechanism description, and source.
The context code and task description follow
(Listing~\ref{lst:knowledge-prompt}).
The no-knowledge condition omits the documentation block;
retrieval replaces it with five retrieved chunks.
C2 uses the shorter prompt in Listing~\ref{lst:c2-prompt},
which omits the self-contained-completion instruction.

\begin{lstlisting}[
  caption={Evaluation prompt. Only the selected knowledge
  components appear in the documentation block.},
  label={lst:knowledge-prompt},
  basicstyle=\ttfamily\footnotesize,
  breaklines=true,
  columns=fullflexible,
  keepspaces=true,
  showstringspaces=false
]
# Reference Documentation:
API: {api_name}

Parameters:
  - {name} ({type}), default={default}: {description}
  ...

Example 1:
```python
{example_code}
```
...

Conceptual background:
{mechanism_description}

Implementation source:
```python
{source_code}
```

---

{context_code}

# Task:
{description}

# Output a self-contained completion for the missing region. If `context_code`
# does not already provide a required import, define a needed name, or set up
# required state, include the missing lines in your completion.
#
# Complete the following code (output only the missing lines, no explanation):
\end{lstlisting}

\subsection{Retrieval Setup}
\label{app:setup-retrieval}

RQ2 replaces the target API's bundle with retrieved documentation.
Each domain's retrieval pool contains Stage-2 bundles from its
benchmark libraries, including APIs without tasks.
The pools contain $130$--$668$ bundles each and $1{,}885$ in total;
the eight \texttt{httpx} bundles are excluded because the library
contributes no benchmark tasks.

For each knowledge condition, bundles are rendered using only
the selected components, split on whitespace, and cut into
windows of $512$ words that start every $448$ words
($64$-word overlap).

We embed chunks with \texttt{BAAI/bge-small-en-v1.5}~\citep{bge}
(CLS pooling, L2-normalized) and build an exact
inner-product FAISS index for each domain and condition.
The query is the task description followed by the context code,
encoded without the model's retrieval instruction.
The encoder truncates chunks and queries to its $512$-token
limit, so the embedding of a longer chunk reflects only its
beginning; this applies to $50\%$ of chunks overall
($81\%$ under Full) and to $6\%$ of queries.
The five highest-scoring chunks, untruncated, replace the oracle
documentation in rank order;
prompting, decoding, and scoring otherwise remain unchanged.
Appendix~\ref{app:rq2-recall} reports target-API retrieval recall.

\subsection{Parametric Adaptation}
\label{app:setup-rq3}

\paragraph{Data split.}
We split the Qwen2.5-Coder-7B task pool $80/20$ by target API,
with no API overlap between training and testing.
Before full-harness filtering, the split contains $1{,}502$
training tasks for $722$ APIs and $397$ test tasks for $190$ APIs.
All methods use the complete training split, including $180$
tasks excluded by full-harness filtering whose reference
solutions remain valid.
Evaluation uses the $348$ test tasks covering $178$ APIs
that belong to the final benchmark.
Each method uses one training run with a fixed seed;
reported confidence intervals capture test-set sampling
uncertainty, not variation across training runs.

\paragraph{Training examples.}
All methods use the same reference-solution targets, containing
the context code and the missing lines.
SFT pairs these targets with the no-knowledge task prompt,
using the model's chat template and a loss over the full
sequence, including the prompt.
AlphaEdit-LoRA uses the same format and loss.
Evaluation requests only the missing lines.

RAFT~\citep{zhang2024raft} adds the target API's Full bundle
and those of two other training APIs in random order.
The target bundle is omitted with probability $0.2$.
MEMIT's edit prompts identify the target in an
\texttt{API: <name>} header; GRACE uses the no-knowledge prompt.
At evaluation, all methods receive the same prompts for
each knowledge condition.

\paragraph{Adaptation methods.}
Table~\ref{tab:hparams} lists the hyperparameters.
Layer indices refer to the $28$ decoder blocks of
Qwen2.5-Coder-7B.
MEMIT's second-moment statistics are estimated from
$20{,}000$ Wikitext samples in FP32 and reused by AlphaEdit-LoRA.
GRACE stores its codebook at the input of layer $24$'s
\texttt{up\_proj}, using cold-initialized values and
coverage-based radius expansion.

Our AlphaEdit-LoRA variant adapts the null-space projection
of AlphaEdit~\citep{alphaedit} to gradient-based LoRA training.
Adapters are attached to \texttt{down\_proj} in layers $5$--$9$.
Before training and after each optimizer step, the adapter input
matrix $A$ is replaced by $AP$, where $P$ projects onto covariance
directions with singular values below $2\times10^{-2}$.
This replaces AlphaEdit's constrained closed-form update with
task-level fine-tuning; we therefore report it separately
as AlphaEdit-LoRA.

\begin{table}[h]
\centering
\small
\caption{Adaptation hyperparameters.
All LoRA adapters use rank $64$.
Batch sizes are per step, before gradient accumulation.}
\label{tab:hparams}
\setlength{\tabcolsep}{4pt}
\begin{adjustbox}{max width=\linewidth}
\begin{tabular}{
  >{\raggedright\arraybackslash}p{1.7cm}
  >{\raggedright\arraybackslash}p{4.0cm}
  >{\raggedright\arraybackslash}p{4.7cm}
  >{\raggedright\arraybackslash}p{3.1cm}}
\toprule
Method & Updated components & Optimization & Other settings \\
\midrule
SFT, RAFT
&
LoRA on q, k, v, o, gate, up, and down projections
in all layers; $\alpha=128$, dropout $0.05$
&
AdamW; lr $2\times10^{-5}$; cosine schedule;
$5\%$ warm-up; $3$ epochs;
batch $1$, accumulation $16$
&
Sequence length $2{,}048$ (SFT) and $16{,}384$ (RAFT); BF16;
gradient checkpointing
\\
\midrule
AlphaEdit-LoRA
&
LoRA on \texttt{down\_proj}, layers $5$--$9$;
$\alpha=32$, dropout $0.1$
&
AdamW; lr $10^{-4}$; cosine schedule;
$5\%$ warm-up; $3$ epochs;
batch $1$, accumulation $16$
&
Sequence length $2{,}048$;
projection of $A$ after each step, threshold $2\times10^{-2}$
\\
\midrule
MEMIT
&
Closed-form updates to \texttt{down\_proj},
layers $5$--$9$
&
$25$ value-optimization steps at lr $0.5$
on layer $27$; weight decay $0.5$;
KL weight $0.0625$; clamp factor $0.75$
&
Subject token \texttt{subject\_last};
covariance weight $15{,}000$
\\
\midrule
GRACE
&
Codebook at layer $24$ \texttt{up\_proj}
&
Adam; lr $1.0$; up to $100$ steps per edit;
early stopping below loss $0.01$
&
Initial radius $\epsilon=1.0$;
one edit per training task
\\
\bottomrule
\end{tabular}
\end{adjustbox}
\end{table}

Inference and adaptation use NVIDIA H100, A100, and L40S GPUs;
sandboxed code evaluation runs on CPU nodes.

\subsection{Statistical Procedures}
\label{app:setup-stats}

\paragraph{Confidence intervals.}
Unless otherwise specified, we report $95\%$ bias-corrected
and accelerated (BCa) bootstrap intervals from $10{,}000$
resamples of target APIs (seed $20260912$).
Tasks sharing an API are resampled together.
For comparisons between conditions, the same resampled tasks
are used to compute paired differences.
Exceptions are the Wilson intervals in
Table~\ref{tab:difficulty-structural} and the human-evaluation
agreement statistics.

\paragraph{Significance tests.}
We use exact McNemar tests for paired binary outcomes,
with Holm correction within each comparison family:
the $36$ pairwise contrasts specified in
Appendix~\ref{app:knowledge-conditions},
the $12$ oracle-versus-retrieval contrasts,
the five RQ3 comparisons against Base, and the per-backbone
families in Appendices~\ref{app:novelty-type}
and~\ref{app:cross-backbone}.
Significance denotes an adjusted $p<0.05$.

\paragraph{API-level robustness check.}
Because McNemar tests operate on individual tasks, we also
test differences at the API level.
We sum paired task differences within each API and apply
a sign-flip permutation test with $10{,}000$ permutations
and the same Holm corrections.
The two procedures agree on all significance decisions except the comparison marked with $\dagger$ in Table~\ref{tab:scale}.

%% file: app/e_rq1.tex
\section{Additional Results for RQ1}
\label{app:rq1-additional}

\subsection{Full Knowledge-Condition Sweep}
\label{app:full-conditions}

Figure~\ref{fig:rq1-components} in the main text reports all twelve
knowledge conditions on Qwen2.5-Coder-7B; this section adds the
comparisons behind it.
Providing the API name alone yields $20.4\%$ Pass@1.
Adding parameter information raises this to $43.0\%$ and
reduces parameter errors from $19.4\%$ to $6.6\%$ of tasks.
Examples provide most of the remaining benefit:
$S{+}E{+}C$ reaches $71.1\%$, close to Full at $71.4\%$.

Of the $36$ pairwise comparisons defined in
Appendix~\ref{app:knowledge-conditions}, $34$ are significant
after Holm correction.
The exceptions are $C$ versus $S{+}M$ ($+3.1$ points,
adjusted $p=0.066$) and Full versus $S{+}E$ ($+1.9$ points,
adjusted $p=0.077$).
Adding $S$ to $E$ gives a modest but significant gain
of $2.3$ points (adjusted $p=0.027$).

\subsection{Novelty Type Analysis}
\label{app:novelty-type}

We compare newly introduced and signature-modified APIs,
as defined in Appendix~\ref{app:bench-pipeline}.
Across all six backbones, examples outperform surface information
much more on newly introduced APIs.
Conversely, adding surface information to examples helps more
on signature-modified APIs (Table~\ref{tab:novelty-cells}).
Both differences are significant on every backbone
(Table~\ref{tab:novelty-tests}), supporting the complementary
roles of usage examples and interface details.

Confidence intervals use API-cluster bootstrap resampling
stratified by novelty type.
Interaction tests permute novelty-type labels across APIs,
with Holm correction across the six backbones for each test.

\begin{table}[h]
\centering
\small
\caption{Pass@1 (\%) by novelty type on each backbone's instance.
$n$ is the number of tasks.
The last column gives the paired difference in percentage points
with its $95\%$ confidence interval.}
\label{tab:novelty-cells}
\setlength{\tabcolsep}{4pt}
\begin{adjustbox}{max width=\linewidth}
\begin{tabular}{llrrrrrc}
\toprule
Backbone & Type & $n$ & $S$ & $E$ & $S{+}E$ & Full & $E-S$ \\
\midrule
Qwen-7B & New
& $1{,}376$ & $39.6$ & $68.5$ & $69.1$ & $70.6$
& $+28.9\;[+25.4, +32.2]$ \\
& Modified
& $294$ & $58.8$ & $61.2$ & $71.4$ & $74.8$
& $+2.4\;[-4.3, +8.7]$ \\
\midrule
Seed-8B & New
& $1{,}368$ & $47.5$ & $69.1$ & $74.8$ & $77.0$
& $+21.6\;[+18.4, +24.8]$ \\
& Modified
& $293$ & $64.5$ & $66.9$ & $79.9$ & $76.1$
& $+2.4\;[-4.8, +9.1]$ \\
\midrule
OpenCoder-8B & New
& $1{,}370$ & $38.3$ & $60.7$ & $60.3$ & $59.4$
& $+22.4\;[+19.2, +25.6]$ \\
& Modified
& $321$ & $49.2$ & $55.5$ & $66.7$ & $61.1$
& $+6.2\;[-0.6, +12.9]$ \\
\midrule
R1-Distill-7B & New
& $1{,}412$ & $14.3$ & $33.6$ & $38.0$ & $38.5$
& $+19.3\;[+16.4, +22.2]$ \\
& Modified
& $328$ & $31.7$ & $32.9$ & $46.0$ & $44.5$
& $+1.2\;[-5.7, +7.9]$ \\
\midrule
Qwen-14B & New
& $1{,}361$ & $54.7$ & $81.0$ & $83.1$ & $83.8$
& $+26.3\;[+23.2, +29.5]$ \\
& Modified
& $267$ & $72.7$ & $73.4$ & $81.6$ & $85.8$
& $+0.7\;[-5.3, +7.0]$ \\
\midrule
Qwen-32B & New
& $1{,}314$ & $57.3$ & $80.6$ & $83.3$ & $84.6$
& $+23.3\;[+20.0, +26.7]$ \\
& Modified
& $255$ & $69.8$ & $76.9$ & $85.1$ & $85.9$
& $+7.1\;[+0.4, +13.5]$ \\
\bottomrule
\end{tabular}
\end{adjustbox}
\end{table}

\begin{table}[h]
\centering
\small
\caption{Differences in knowledge gains between novelty types,
in percentage points with $95\%$ confidence intervals.
Left: the greater benefit of $E$ over $S$ on new APIs.
Right: the greater benefit of adding $S$ to $E$ on modified APIs.
$p$-values are Holm-adjusted across backbones within each test.}
\label{tab:novelty-tests}
\setlength{\tabcolsep}{4pt}
\begin{adjustbox}{max width=\linewidth}
\begin{tabular}{lcccc}
\toprule
& \multicolumn{2}{c}{$(E-S)_{\mathrm{new}}-(E-S)_{\mathrm{mod}}$}
& \multicolumn{2}{c}{$(SE-E)_{\mathrm{mod}}-(SE-E)_{\mathrm{new}}$} \\
\cmidrule(lr){2-3}
\cmidrule(lr){4-5}
Backbone & Difference & Adj.\ $p$ & Difference & Adj.\ $p$ \\
\midrule
Qwen-7B
& $+26.5\;[+19.3, +33.7]$ & $0.0012$
& $+9.6\;[+4.8, +14.4]$ & $0.0012$ \\
Seed-8B
& $+19.2\;[+11.6, +27.0]$ & $0.0012$
& $+7.3\;[+1.7, +13.0]$ & $0.0078$ \\
OpenCoder-8B
& $+16.2\;[+8.8, +23.8]$ & $0.0012$
& $+11.7\;[+5.9, +17.4]$ & $0.0012$ \\
R1-Distill-7B
& $+18.0\;[+10.6, +25.4]$ & $0.0012$
& $+8.7\;[+2.6, +14.8]$ & $0.0084$ \\
Qwen-14B
& $+25.6\;[+18.7, +32.4]$ & $0.0012$
& $+6.1\;[+1.5, +10.8]$ & $0.0016$ \\
Qwen-32B
& $+16.2\;[+9.0, +23.4]$ & $0.0012$
& $+5.5\;[+0.5, +10.7]$ & $0.0084$ \\
\bottomrule
\end{tabular}
\end{adjustbox}
\end{table}

\subsection{Cross-Backbone and Scale Results}
\label{app:cross-backbone}

Table~\ref{tab:cross-model} reports results on each backbone's
model-conditioned instance.
Across all six models, examples outperform surface information,
and $S{+}E$ performs close to Full.
For direct comparisons between backbones,
Figure~\ref{fig:rq1-novelty-backbones}b uses the $1{,}405$
tasks shared by all six instances.
On this common set, $E-S$ ranges from $17.9$ to $25.3$
percentage points, while Full differs from $S{+}E$ by
only $-1.5$ to $+1.8$ points.

\begin{table}[t]
\centering
\small
\caption{Pass@1 (\%) on each backbone's own instance,
with $95\%$ API-cluster bootstrap confidence intervals.
Task counts differ because C2 filters each instance separately.}
\label{tab:cross-model}
\setlength{\tabcolsep}{3pt}
\begin{adjustbox}{max width=\linewidth}
\begin{tabular}{lcccccc}
\toprule
Condition & Qwen-7B & Qwen-14B & Qwen-32B
& OpenCoder-8B & Seed-8B & R1-Distill-7B \\
\midrule
Tasks & $1{,}670$ & $1{,}628$ & $1{,}569$
& $1{,}691$ & $1{,}661$ & $1{,}740$ \\
\midrule
No knowledge
& $0.8\;[0.4, 1.3]$
& $0.7\;[0.4, 1.3]$
& $0.6\;[0.3, 1.0]$
& $1.7\;[1.1, 2.5]$
& $1.1\;[0.7, 1.8]$
& $0.7\;[0.4, 1.3]$ \\
$S$
& $43.0\;[40.2, 45.8]$
& $57.6\;[54.7, 60.4]$
& $59.3\;[56.3, 62.2]$
& $40.4\;[37.8, 43.0]$
& $50.5\;[47.6, 53.3]$
& $17.6\;[15.7, 19.6]$ \\
$E$
& $67.2\;[64.8, 69.6]$
& $79.7\;[77.5, 81.8]$
& $80.0\;[77.6, 82.2]$
& $59.7\;[57.2, 62.2]$
& $68.7\;[66.2, 71.1]$
& $33.4\;[31.0, 35.9]$ \\
$M$
& $3.2\;[2.4, 4.4]$
& $4.4\;[3.4, 5.8]$
& $4.3\;[3.1, 5.8]$
& $3.4\;[2.6, 4.5]$
& $2.6\;[1.7, 3.7]$
& $1.3\;[0.8, 1.9]$ \\
$C$
& $50.0\;[47.2, 52.7]$
& $56.4\;[53.5, 59.3]$
& $61.8\;[58.9, 64.7]$
& $39.7\;[37.2, 42.2]$
& $52.1\;[49.3, 54.8]$
& $13.6\;[12.0, 15.4]$ \\
$S{+}M$
& $46.9\;[44.0, 49.7]$
& $60.1\;[57.2, 62.9]$
& $59.6\;[56.4, 62.6]$
& $40.5\;[37.9, 43.2]$
& $51.6\;[48.7, 54.4]$
& $22.8\;[20.8, 25.0]$ \\
$S{+}E$
& $69.5\;[67.2, 71.8]$
& $82.9\;[80.8, 84.7]$
& $83.6\;[81.3, 85.7]$
& $61.5\;[59.1, 63.9]$
& $75.7\;[73.3, 77.9]$
& $39.5\;[36.9, 42.1]$ \\
$S{+}C$
& $58.1\;[55.5, 60.9]$
& $70.2\;[67.6, 72.8]$
& $71.3\;[68.6, 74.0]$
& $49.7\;[47.2, 52.4]$
& $63.3\;[60.7, 66.0]$
& $23.4\;[21.3, 25.6]$ \\
Full
& $71.4\;[69.0, 73.6]$
& $84.2\;[82.3, 86.0]$
& $84.8\;[82.7, 86.7]$
& $59.7\;[57.4, 62.1]$
& $76.9\;[74.6, 79.0]$
& $39.7\;[37.2, 42.2]$ \\
\bottomrule
\end{tabular}
\end{adjustbox}
\end{table}

\paragraph{Model scale.}
On the $1{,}476$ tasks shared by the three Qwen2.5-Coder
instances, performance improves substantially from 7B to 14B,
with smaller changes at 32B (Table~\ref{tab:scale}).
Adding examples to $S$ yields gains of $24$--$28$ points
at every size, whereas the gain from adding $M$ declines
from $3.3$ points at 7B to $0.3$ at 32B.
Full adds only $1.2$--$1.6$ points over $S{+}E$.

\begin{table}[h]
\centering
\small
\caption{Pass@1 (\%) on $1{,}476$ shared Qwen2.5-Coder tasks.
The lower rows report paired differences in percentage points;
brackets give $95\%$ API-cluster bootstrap confidence intervals.
$^{*}$ indicates Holm-adjusted McNemar significance within
each size; $^{\dagger}$ indicates that significance does not
hold under the API-level robustness check
(Appendix~\ref{app:setup-stats}).}
\label{tab:scale}
\setlength{\tabcolsep}{4pt}
\begin{tabular}{lccc}
\toprule
Condition or contrast & 7B & 14B & 32B \\
\midrule
$S$
& $42.8\;[39.7, 45.8]$
& $57.8\;[54.7, 60.8]$
& $59.9\;[56.8, 62.8]$ \\
$E$
& $68.2\;[65.6, 70.7]$
& $80.9\;[78.6, 83.0]$
& $80.3\;[77.9, 82.6]$ \\
$S{+}M$
& $46.1\;[43.0, 49.1]$
& $60.3\;[57.2, 63.3]$
& $60.2\;[57.0, 63.3]$ \\
$S{+}E$
& $70.5\;[67.8, 72.8]$
& $83.4\;[81.3, 85.4]$
& $84.0\;[81.6, 86.1]$ \\
Full
& $71.6\;[69.1, 74.0]$
& $85.0\;[83.1, 86.9]$
& $85.4\;[83.3, 87.3]$ \\
\midrule
$(S{+}M)-S$
& $+3.3\;[+1.2, +5.4]^{*}$
& $+2.5\;[+0.4, +4.5]^{*}$
& $+0.3\;[-1.8, +2.5]$ \\
$(S{+}E)-S$
& $+27.7\;[+24.6, +30.8]^{*}$
& $+25.6\;[+22.7, +28.5]^{*}$
& $+24.1\;[+21.3, +27.1]^{*}$ \\
$\mathrm{Full}-(S{+}E)$
& $+1.2\;[-1.1, +3.4]$
& $+1.6\;[+0.0, +3.3]^{*\dagger}$
& $+1.4\;[-0.5, +3.2]$ \\
\bottomrule
\end{tabular}
\end{table}

\subsection{Results by Domain and Library}
\label{app:domain}

Examples outperform surface information in all five domains
(Table~\ref{tab:domain}).
The smaller gain in data science is consistent with its higher
share of signature-modified APIs, for which $E$ and $S$
perform similarly.
Restricting the comparison to newly introduced APIs yields
gains of $26$--$39$ percentage points across domains.
Evidence for between-domain differences does not survive
Holm correction (adjusted $p=0.20$).

At the library level, $E-S$ is positive in $18$ of $20$
libraries, with $95\%$ intervals excluding zero in nine.
The two negative estimates occur in \texttt{scanpy} and
\texttt{flask}, represented by only seven and three tasks,
respectively.

\begin{table}[h]
\centering
\small
\caption{Pass@1 (\%) by domain on Qwen2.5-Coder-7B.
The final row gives the paired $E-S$ difference in percentage
points with $95\%$ API-cluster bootstrap confidence intervals.}
\label{tab:domain}
\setlength{\tabcolsep}{4pt}
\begin{adjustbox}{max width=\linewidth}
\begin{tabular}{lccccc}
\toprule
Condition & Agent tools & AI for science & Data science
& Deep learning & Software eng. \\
\midrule
Tasks & $268$ & $390$ & $238$ & $700$ & $74$ \\
\midrule
No knowledge & $0.7$ & $1.5$ & $0.8$ & $0.4$ & $0.0$ \\
$S$       & $37.7$ & $35.6$ & $47.5$ & $48.3$ & $36.5$ \\
$E$       & $64.6$ & $64.4$ & $59.2$ & $72.7$ & $64.9$ \\
$M$       & $1.1$  & $4.4$  & $6.7$  & $1.9$  & $6.8$ \\
$C$       & $39.6$ & $47.4$ & $49.6$ & $56.4$ & $41.9$ \\
$S{+}M$   & $38.1$ & $40.3$ & $51.3$ & $52.6$ & $45.9$ \\
$S{+}E$   & $65.3$ & $66.4$ & $66.4$ & $73.9$ & $70.3$ \\
$S{+}C$   & $47.8$ & $52.8$ & $58.4$ & $66.0$ & $48.6$ \\
Full      & $66.0$ & $65.9$ & $74.4$ & $74.9$ & $77.0$ \\
\midrule
$E-S$
& \shortstack{$+26.9$ \\ $[+19.2, +34.8]$}
& \shortstack{$+28.7$ \\ $[+22.2, +35.1]$}
& \shortstack{$+11.8$ \\ $[+3.2, +20.6]$}
& \shortstack{$+24.4$ \\ $[+20.0, +28.9]$}
& \shortstack{$+28.4$ \\ $[+13.1, +43.5]$} \\
\bottomrule
\end{tabular}
\end{adjustbox}
\end{table}

\subsection{Results by Difficulty}
\label{app:difficulty}

\begin{figure}[h]
\centering
\includegraphics[width=\linewidth]{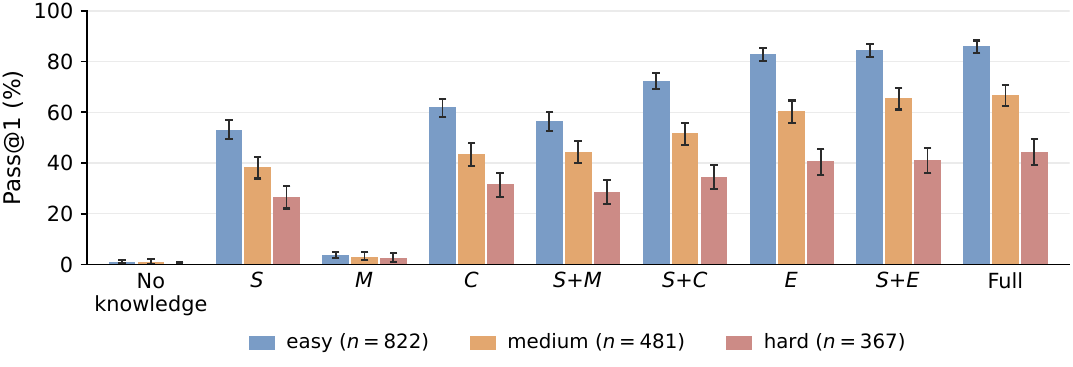}
\caption{Pass@1 by difficulty tier on Qwen2.5-Coder-7B,
with $95\%$ API-cluster bootstrap confidence intervals.
The tiers contain $822$ easy, $481$ medium, and $367$ hard tasks.}
\label{fig:difficulty}
\end{figure}

Performance decreases from easy to hard under every knowledge
condition (Figure~\ref{fig:difficulty}).
Even with the Full bundle, Pass@1 falls from $86.0\%$ on easy
tasks to $66.9\%$ on medium and $44.4\%$ on hard tasks.
Examples remain more effective than surface information
at every tier, though the $E-S$ gap narrows from $29.8$
to $14.2$ percentage points.
These results are consistent with the increasing integration
demands of the tiers: supplying API knowledge does not
eliminate the difficulty of using it in more complex code.

\subsection{Pass@5 and Auxiliary Metrics}
\label{app:aux-metrics}

Table~\ref{tab:aux-metrics} reports Pass@5 from $20$ samples
at $T=0.8$, alongside two diagnostics of greedy completions:
the proportion containing a target-API call and Pass@1
within that subset.
Calls are identified from the syntax tree, excluding
completions that define a replacement function with the
target's name; this measures call presence in code,
not whether the call executes.

Pass@5 preserves the ordering of conditions, with $E$ reaching
$82.0\%$ compared with $61.9\%$ for $S$.
Examples also increase the proportion of completions containing
a target call ($88.2\%$ versus $72.9\%$) and the pass rate
within that subset ($76.2\%$ versus $58.9\%$).
Thus, their advantage appears in both target-API selection
and successful use.

\begin{table}[h]
\centering
\small
\caption{Auxiliary metrics (\%) on Qwen2.5-Coder-7B
($1{,}670$ tasks).
Call presence is detected statically in greedy completions;
conditional Pass@1 is computed on tasks with a detected call.
Brackets show $95\%$ API-cluster bootstrap confidence intervals.}
\label{tab:aux-metrics}
\setlength{\tabcolsep}{4pt}
\begin{adjustbox}{max width=\linewidth}
\begin{tabular}{lrccc}
\toprule
Condition & Pass@1 & Pass@5 & Call presence
& Pass@1 $\mid$ call present \\
\midrule
No knowledge
& $0.8$
& $2.1\;[1.6, 2.8]$
& $5.7\;[4.5, 7.3]$
& $13.5\;[7.3, 22.3]$ \\
$S_{\text{name}}$
& $20.4$
& $31.8\;[29.4, 34.4]$
& $55.7\;[52.6, 58.7]$
& $36.6\;[33.1, 40.2]$ \\
$S$
& $43.0$
& $61.9\;[59.2, 64.3]$
& $72.9\;[70.2, 75.6]$
& $58.9\;[55.8, 62.1]$ \\
$E$
& $67.2$
& $82.0\;[80.2, 83.8]$
& $88.2\;[86.3, 89.9]$
& $76.2\;[73.8, 78.4]$ \\
$M$
& $3.2$
& $6.1\;[5.0, 7.4]$
& $14.0\;[12.0, 16.2]$
& $23.2\;[17.6, 30.0]$ \\
$C$
& $50.0$
& $70.4\;[68.0, 72.6]$
& $80.2\;[77.9, 82.4]$
& $62.3\;[59.4, 65.2]$ \\
$S{+}M$
& $46.9$
& $64.7\;[62.1, 67.1]$
& $75.3\;[72.7, 77.9]$
& $62.2\;[59.1, 65.2]$ \\
$S{+}E$
& $69.5$
& $84.9\;[83.2, 86.4]$
& $89.3\;[87.5, 90.8]$
& $77.9\;[75.5, 80.1]$ \\
$S{+}C$
& $58.1$
& $76.5\;[74.2, 78.5]$
& $85.5\;[83.4, 87.4]$
& $68.0\;[65.2, 70.6]$ \\
Full
& $71.4$
& $86.0\;[84.4, 87.5]$
& $89.6\;[87.8, 91.1]$
& $79.7\;[77.5, 81.7]$ \\
\bottomrule
\end{tabular}
\end{adjustbox}
\end{table}

\subsection{Failure Composition under Sampling}
\label{app:pass5-taxonomy}

We compare greedy decoding with sampling at $T=0.8$ under
five knowledge conditions: no knowledge, $S$, $E$, $S{+}E$,
and Full.
For each of the $1{,}670$ tasks, we analyze the first five
of the twenty completions used to estimate Pass@5,
using the same harness and failure classifier as in the
main analysis.

\begin{figure}[h]
\centering
\includegraphics[width=\linewidth]{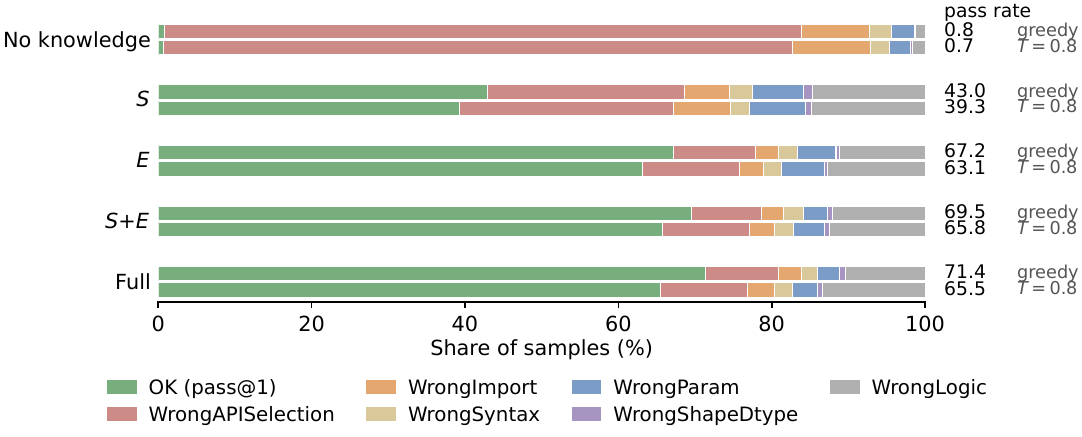}
\caption{Outcome composition under greedy decoding and sampling
on Qwen2.5-Coder-7B ($1{,}670$ tasks per condition).
Sampling rows aggregate five completions per task at $T=0.8$.
The right column reports per-completion pass rates, not Pass@5.}
\label{fig:pass5-taxonomy}
\end{figure}

\paragraph{Patterns across knowledge conditions.}
Sampling produces lower per-completion pass rates than greedy
decoding, but preserves the performance ranking and the main
failure patterns (Figure~\ref{fig:pass5-taxonomy}).
Target-API selection errors account for $81.9\%$ of samples
without knowledge, falling to $27.9\%$ with $S$ and
$11.3$--$12.7\%$ with example-containing bundles.
The reduction in selection errors observed under greedy
decoding therefore also holds under sampling.

\paragraph{Consistency within tasks.}
Among tasks with at least one failed sample, $38$--$53\%$
exhibit multiple failure categories across their five samples.
Randomly redistributing failure labels within each condition,
while preserving each task's number of failures, raises this
proportion to $61$--$80\%$ across $1{,}000$ permutations.
Failures thus tend to recur within a task, although their
categories can vary between completions.
The taxonomy describes individual completions instead of
assigning a fixed failure type to each task.

\subsection{Sensitivity to Construction Examples}
\label{app:fresh-examples}

Stage~3 generates tasks from API knowledge bundles, including their usage examples, and C3 retains tasks solvable by GPT-5-mini with the Full bundle (Appendix~\ref{app:bench-pipeline}). This may favor the examples used during construction. To assess sensitivity to this choice, we generate fresh examples for the $856$ target APIs in the Qwen2.5-Coder-7B instance using the same Stage-2 prompt, model, and validation procedure. Neither task generation nor filtering uses these fresh examples. For $25$ APIs, no fresh example passes validation; excluding their $53$ tasks leaves $1{,}617$ tasks covering $831$ APIs.

Both example sets contain an average of $2.5$ examples per API. Only one of the $2{,}068$ fresh examples duplicates an original example. The fresh set has a slightly higher proportion of static rather than successfully executed examples ($14.6\%$ versus $11.8\%$).

We evaluate $S$ and both versions of $E$, $S{+}E$, and Full in the same inference run using the main evaluation harness. In $S{+}E$ and Full, only the examples are replaced. Rerunning the original-example conditions provides a baseline under the same inference setup; their task outcomes agree with the main run on $98\%$ of tasks.

\paragraph{Overlap with reference completions.}
Original examples have greater overlap with reference completions than fresh examples do. In $85.3\%$ of tasks, the reference call uses the same set of keyword argument names as at least one original example, compared with $68.2\%$ for fresh examples. The mean fraction of reference code-token $4$-grams also present in an example decreases from $0.28$ to $0.19$. 

\paragraph{Results.}
Replacing the original examples lowers Pass@1 by $6.9$ percentage points under $E$, $4.3$ under $S{+}E$, and $5.4$ under Full (Table~\ref{tab:fresh-examples}; exact McNemar $p<10^{-4}$ for each comparison). Nevertheless, \textbf{fresh examples remain the strongest standalone component overall}, exceeding $S$ by $16.5$ points and $C$ by $10.6$ points. Full provides little additional gain over $S{+}E$.

The results also reinforce the importance of interface information for signature-modified APIs. Fresh examples outperform $S$ on newly introduced APIs, but fall $8.0$ points below $S$ on signature-modified APIs, where adding $S$ to $E$ improves Pass@1 by $15.0$ points.

The performance decrease is consistent with an advantage from alignment with the construction examples, although replacing examples also changes their content and coverage. 

\begin{table}[h]
\centering
\small
\caption{Original versus fresh examples on the same $1{,}617$ Qwen2.5-Coder-7B tasks. The first three rows report Pass@1 (\%); subsequent rows report paired differences in percentage points. Brackets give $95\%$ API-cluster bootstrap confidence intervals. Differences are computed before rounding. $S$ ($44.5\%$) is rerun in the same inference run; $C$ ($50.3\%$) is taken from the main run on the same task subset. In $S{+}E$ and Full, only the example component is replaced.}
\label{tab:fresh-examples}
\setlength{\tabcolsep}{4pt}
\begin{adjustbox}{max width=\linewidth}
\begin{tabular}{lccc}
\toprule
& Original examples & New examples & New $-$ Original \\
\midrule
$E$ & $67.8$ & $61.0$ & $-6.9\;[-9.2, -4.6]$ \\
$S{+}E$ & $69.6$ & $65.4$ & $-4.3\;[-6.4, -2.1]$ \\
Full & $70.9$ & $65.5$ & $-5.4\;[-7.4, -3.4]$ \\
\midrule
$E-S$ & $+23.4\;[+20.2, +26.4]$ & $+16.5\;[+13.6, +19.7]$ & \\
$E-C$ & $+17.5\;[+14.2, +20.6]$ & $+10.6\;[+7.4, +13.8]$ & \\
Full $-\,(S{+}E)$ & $+1.3\;[-1.0, +3.4]$ & $+0.1\;[-2.0, +2.4]$ & \\
\midrule
\multicolumn{4}{l}{\emph{Newly introduced APIs} ($n=1{,}330$)} \\
$E-S$ & $+28.1\;[+24.6, +31.4]$ & $+21.8\;[+18.5, +25.2]$ & \\
$(S{+}E)-E$ & $+0.3\;[-1.6, +2.2]$ & $+2.1\;[+0.2, +4.1]$ & \\
\multicolumn{4}{l}{\emph{Signature-modified APIs} ($n=287$)} \\
$E-S$ & $+1.4\;[-4.8, +7.7]$ & $-8.0\;[-14.2, -1.8]$ & \\
$(S{+}E)-E$ & $+8.7\;[+4.4, +13.5]$ & $+15.0\;[+10.2, +20.2]$ & \\
\bottomrule
\end{tabular}
\end{adjustbox}
\end{table}

%% file: app/f_rq2.tex
\section{Additional Results for RQ2}
\label{app:rq2-additional}

\subsection{Oracle vs.\ Retrieved Knowledge}
\label{app:rq2-all-conditions}

Retrieval reduces Pass@1 in all eleven knowledge-bearing
conditions (Table~\ref{tab:rq2-all}).
All reductions are significant after Holm correction.
Full and $S{+}E$ remain the strongest retrieved conditions,
at $51.0\%$ and $49.4\%$, respectively.
Examples alone reach $44.8\%$, well above surface information
($25.6\%$) and implementation source ($26.1\%$).
Thus, the advantage of examples persists under retrieval,
despite lower overall performance.

\begin{table}[h]
\centering
\small
\caption{Pass@1 (\%) with oracle and retrieved knowledge on
Qwen2.5-Coder-7B ($1{,}670$ tasks).
Paired differences are in percentage points with $95\%$
API-cluster bootstrap confidence intervals.
$p$-values are Holm-adjusted McNemar results.}
\label{tab:rq2-all}
\setlength{\tabcolsep}{4pt}
\begin{tabular}{lrrcc}
\toprule
Condition & Oracle & Retrieved & Retrieved $-$ oracle & Adj.\ $p$ \\
\midrule
No knowledge
& $0.8$ & $0.7$ & $-0.1\;[-0.3, +0.0]$ & $1$ \\
$S_{\text{name}}$
& $20.4$ & $8.1$ & $-12.3\;[-14.4, -10.2]$ & $<10^{-3}$ \\
$S$
& $43.0$ & $25.6$ & $-17.4\;[-20.2, -14.7]$ & $<10^{-3}$ \\
$E$
& $67.2$ & $44.8$ & $-22.4\;[-24.9, -19.8]$ & $<10^{-3}$ \\
$M$
& $3.2$ & $2.0$ & $-1.3\;[-2.3, -0.4]$ & $0.006$ \\
$C$
& $50.0$ & $26.1$ & $-23.9\;[-26.7, -21.2]$ & $<10^{-3}$ \\
$S{+}M$
& $46.9$ & $34.7$ & $-12.2\;[-14.9, -9.5]$ & $<10^{-3}$ \\
$S{+}E$
& $69.5$ & $49.4$ & $-20.1\;[-22.7, -17.6]$ & $<10^{-3}$ \\
$S{+}C$
& $58.1$ & $35.9$ & $-22.2\;[-25.0, -19.5]$ & $<10^{-3}$ \\
$S{+}M{+}C$
& $60.7$ & $41.0$ & $-19.7\;[-22.4, -17.0]$ & $<10^{-3}$ \\
$S{+}E{+}C$
& $71.1$ & $47.7$ & $-23.4\;[-26.0, -20.7]$ & $<10^{-3}$ \\
Full
& $71.4$ & $51.0$ & $-20.4\;[-23.2, -17.7]$ & $<10^{-3}$ \\
\bottomrule
\end{tabular}
\end{table}

\subsection{Target Recall and Conditional Performance}
\label{app:rq2-recall}

A retrieval \emph{hit} contains at least one chunk from the
target API's bundle among the five retrieved chunks.
Recall varies with the indexed components, reaching $77.9\%$
for $E$ and $77.4\%$ for Full, compared with $62.0\%$ for $S$
(Table~\ref{tab:rq2-recall}).

When retrieval misses the target, Pass@1 is at most $1.9\%$
across the reported conditions.
Even on hits, retrieved knowledge generally falls short of
the oracle bundle.
A hit guarantees only partial coverage of the target
documentation; incomplete content, chunk boundaries, and
other APIs' documentation may all contribute to the remaining
gap.

To compare conditions on identical tasks,
Figure~\ref{fig:rq2-retrieval}b uses the $960$ tasks with hits
under $S$, $S{+}M$, $S{+}E$, and Full.
The retrieved-minus-oracle differences are $-3.9$, $+0.1$,
$-6.6$, and $-6.4$ points, respectively.
The gaps for $S{+}E$ and Full therefore persist even
when the target is retrieved under every compared condition.

\begin{table}[h]
\centering
\small
\caption{Target recall and conditional Pass@1 (\%) on
Qwen2.5-Coder-7B ($1{,}670$ tasks).
A hit requires at least one target-API chunk in the top five.
Differences compare retrieved and oracle performance on the
same hit subset, with $95\%$ API-cluster bootstrap intervals.}
\label{tab:rq2-recall}
\setlength{\tabcolsep}{3.5pt}
\begin{adjustbox}{max width=\linewidth}
\begin{tabular}{lrrrccrrr}
\toprule
& & \multicolumn{4}{c}{Target retrieved}
& \multicolumn{3}{c}{Target missed} \\
\cmidrule(lr){3-6}
\cmidrule(lr){7-9}
Condition & Recall & $n$ & Retrieved & Oracle & Difference
& $n$ & Retrieved & Oracle \\
\midrule
$S$
& $62.0$ & $1{,}036$ & $40.6$ & $44.9$
& $-4.2\;[-7.3, -1.3]$
& $634$ & $0.9$ & $39.9$ \\
$E$
& $77.9$ & $1{,}301$ & $57.0$ & $70.3$
& $-13.2\;[-15.8, -10.6]$
& $369$ & $1.6$ & $56.4$ \\
$M$
& $69.7$ & $1{,}164$ & $2.4$ & $2.8$
& $-0.4\;[-1.4, +0.5]$
& $506$ & $1.0$ & $4.2$ \\
$C$
& $58.0$ & $969$ & $43.8$ & $54.8$
& $-11.0\;[-14.1, -7.9]$
& $701$ & $1.7$ & $43.4$ \\
$S{+}M$
& $75.9$ & $1{,}268$ & $45.4$ & $48.3$
& $-2.8\;[-5.5, -0.2]$
& $402$ & $1.0$ & $42.5$ \\
$S{+}E$
& $76.6$ & $1{,}279$ & $64.3$ & $71.6$
& $-7.3\;[-9.8, -4.9]$
& $391$ & $0.8$ & $62.7$ \\
$S{+}C$
& $68.7$ & $1{,}147$ & $51.9$ & $62.1$
& $-10.2\;[-13.0, -7.4]$
& $523$ & $1.0$ & $49.5$ \\
Full
& $77.4$ & $1{,}292$ & $65.3$ & $74.7$
& $-9.4\;[-11.9, -6.8]$
& $378$ & $1.9$ & $60.1$ \\
\bottomrule
\end{tabular}
\end{adjustbox}
\end{table}

%% file: app/g_rq3.tex
\section{Additional Results for RQ3}
\label{app:rq3-additional}

\subsection{Failure Breakdown by Method}
\label{app:rq3-failures}

Without the Full bundle, API-selection errors dominate across
methods (Table~\ref{tab:rq3-failures}).
With the bundle, SFT and RAFT reduce these errors from $9.5\%$
to $1.4\%$ and $2.0\%$, respectively, leaving program logic
as their largest remaining failure category.

Of the base model's $100$ failures with the bundle, SFT solves
$57$ and RAFT solves $69$, including substantial numbers of
selection and logic failures (Table~\ref{tab:rq3-fixes}).
They also fail on $17$ and $20$ previously solved tasks,
yielding net gains of $40$ and $49$ tasks.

\begin{table}[h]
\centering
\small
\caption{Outcome composition (\% of $348$ test tasks).
Failure columns denote API selection, import, syntax,
parameter, shape/dtype, and program-logic errors.}
\label{tab:rq3-failures}
\setlength{\tabcolsep}{4pt}
\begin{adjustbox}{max width=\linewidth}
\begin{tabular}{llrrrrrrr}
\toprule
Method & Bundle & Pass@1 & Sel. & Imp. & Syn. & Par. & S/D & Log. \\
\midrule
Base & None & $0.3$ & $79.0$ & $11.8$ & $2.6$ & $4.6$ & $0.0$ & $1.7$ \\
     & Full & $71.3$ & $9.5$ & $3.7$ & $1.1$ & $3.4$ & $0.3$ & $10.6$ \\
\midrule
SFT & None & $2.3$ & $61.2$ & $25.9$ & $0.0$ & $9.2$ & $0.0$ & $1.4$ \\
    & Full & $82.8$ & $1.4$ & $1.4$ & $0.9$ & $2.9$ & $0.6$ & $10.1$ \\
\midrule
RAFT & None & $1.7$ & $76.4$ & $14.7$ & $0.0$ & $6.3$ & $0.0$ & $0.9$ \\
     & Full & $85.3$ & $2.0$ & $0.6$ & $0.6$ & $2.3$ & $1.4$ & $7.8$ \\
\midrule
GRACE & None & $0.3$ & $79.0$ & $11.8$ & $3.2$ & $4.6$ & $0.0$ & $1.1$ \\
      & Full & $70.1$ & $10.3$ & $4.6$ & $0.9$ & $3.4$ & $0.3$ & $10.3$ \\
\midrule
MEMIT & None & $1.7$ & $77.3$ & $11.8$ & $1.7$ & $6.0$ & $0.0$ & $1.4$ \\
      & Full & $72.1$ & $8.0$ & $3.2$ & $2.6$ & $2.3$ & $0.9$ & $10.9$ \\
\midrule
AlphaEdit-LoRA & None & $1.1$ & $68.4$ & $20.1$ & $1.1$ & $7.5$ & $0.0$ & $1.7$ \\
               & Full & $75.6$ & $6.6$ & $0.6$ & $0.9$ & $2.9$ & $0.0$ & $13.5$ \\
\bottomrule
\end{tabular}
\end{adjustbox}
\end{table}

\begin{table}[h]
\centering
\small
\caption{Base-model failures solved by SFT and RAFT,
with the Full bundle supplied to all models.
The final rows account for previously solved tasks lost
and the resulting net gains.}
\label{tab:rq3-fixes}
\begin{tabular}{lrrr}
\toprule
Base failure category & Tasks & SFT solves & RAFT solves \\
\midrule
API selection & $33$ & $23$ & $26$ \\
Import        & $13$ & $8$  & $9$ \\
Syntax        & $4$  & $1$  & $2$ \\
Parameter     & $12$ & $9$  & $8$ \\
Shape/dtype   & $1$  & $0$  & $0$ \\
Program logic & $37$ & $16$ & $24$ \\
\midrule
All failures       & $100$ & $57$ & $69$ \\
Base passes lost   & $248$ & $17$ & $20$ \\
Net tasks gained   & ---   & $40$ & $49$ \\
\bottomrule
\end{tabular}
\end{table}

\subsection{Bypass Behavior}
\label{app:rq3-bypass}

For completions with no recorded target-API call, we distinguish
four behaviors: implementing the functionality without naming
the target, defining a replacement with the same name,
referencing the target without invoking it, and producing
no extractable code.
We define \emph{bypass} as the first two behaviors.

With the Full bundle, SFT and RAFT reduce bypass from $24$
tasks to one each (Table~\ref{tab:rq3-bypass}).
GRACE remains close to Base, while MEMIT and AlphaEdit-LoRA
also reduce bypass.

Bypass also occurs across all six untuned backbones in the
Full condition.
Rates range from $1.0\%$ for Qwen-32B to $12.1\%$ for
R1-Distill-7B.
Within the Qwen family, the rates are $7.2\%$, $1.4\%$,
and $1.0\%$ for 7B, 14B, and 32B, respectively,
each measured on its own instance.

\begin{table}[h]
\centering
\small
\caption{Completions with no recorded target-API call
($348$ tasks).
Bypass combines reimplementation and same-name redefinition.
$\Delta$ is the paired change in bypass rate from Base
with Full, in percentage points.
Brackets give $95\%$ API-cluster bootstrap confidence intervals.}
\label{tab:rq3-bypass}
\setlength{\tabcolsep}{4pt}
\begin{adjustbox}{max width=\linewidth}
\begin{tabular}{llrrrrcc}
\toprule
Method & Bundle & Reimpl. & Same-name & Not invoked & Empty
& Bypass (\%) & $\Delta$ \\
\midrule
Base & None
& $271$ & $1$ & $0$ & $1$
& $78.2\;[72.8, 82.7]$ & --- \\
Base & Full
& $18$ & $6$ & $6$ & $1$
& $6.9\;[4.4, 10.6]$ & --- \\
SFT & Full
& $1$ & $0$ & $1$ & $1$
& $0.3\;[0.0, 1.6]$ & $-6.6\;[-10.4, -4.0]$ \\
RAFT & Full
& $1$ & $0$ & $1$ & $3$
& $0.3\;[0.0, 1.5]$ & $-6.6\;[-10.4, -4.1]$ \\
GRACE & Full
& $19$ & $6$ & $9$ & $1$
& $7.2\;[4.6, 10.9]$ & $+0.3\;[-0.6, +1.4]$ \\
MEMIT & None
& $261$ & $1$ & $3$ & $2$
& $75.3\;[69.3, 80.3]$ & --- \\
MEMIT & Full
& $9$ & $6$ & $9$ & $1$
& $4.3\;[2.4, 7.3]$ & $-2.6\;[-4.9, -0.9]$ \\
AlphaEdit-LoRA & Full
& $13$ & $0$ & $3$ & $6$
& $3.7\;[1.9, 6.6]$ & $-3.2\;[-6.1, -0.6]$ \\
\bottomrule
\end{tabular}
\end{adjustbox}
\end{table}

\subsection{Paired Code Examples}
\label{app:rq3-examples}

Listing~\ref{lst:rq3-pair} compares Base and SFT on an
authentication-field extraction task, both with the Full bundle.
Base manually removes fields from the request, while SFT
calls \texttt{get\_and\_pop\_keys} and converts its
\texttt{KeyError} into the required \texttt{ValueError}.
The example illustrates both target-API selection and
adaptation of its behavior to the surrounding program.
Additional pairs are provided with the released artifacts.

\begin{lstlisting}[
  caption={Base and SFT completions for the same task,
  both with the Full bundle. Code is abridged.},
  label={lst:rq3-pair},
  basicstyle=\ttfamily\footnotesize,
  breaklines=true,
  columns=fullflexible,
  keepspaces=true,
  showstringspaces=false
]
# Base: manual implementation; fails the target-call check.
auth_info, missing_keys = {}, []
for field in ['auth_token', 'user_id']:
    if field in incoming_request:
        auth_info[field] = incoming_request.pop(field)
    else:
        missing_keys.append(field)
if missing_keys:
    raise ValueError(f"Missing required auth fields: {', '.join(missing_keys)}")

# SFT: uses the target API; passes.
from deepchem.utils.differentiation_utils.misc import get_and_pop_keys
try:
    auth_info = get_and_pop_keys(
        incoming_request, ['auth_token', 'user_id', 'session_id'])
except KeyError as e:
    raise ValueError(f"Missing required auth fields: {e}") from e
\end{lstlisting}

\subsection{Exposure to the Target API}
\label{app:same-api}

The API-disjoint setting in RQ3 tests transfer to APIs absent
from adaptation training.
Here we ask whether training on the target API itself improves
performance on new tasks using that API.

\paragraph{Setup.}
We construct three folds from $296$ APIs with one task per
difficulty tier before full-harness filtering.
Each fold holds out one tier and trains on the other two,
giving $592$ same-API training tasks.
A cross-API control uses the same number of tasks from other
APIs, matched by library.
Evaluation pools the $780$ held-out tasks covering $289$ APIs
that remain in the final benchmark.

We compare the standard SFT configuration
(Appendix~\ref{app:setup-rq3}) with an extended schedule of
$10$ epochs at learning rate $10^{-4}$, using three seeds
for the extended schedule.

\paragraph{Results.}
Under the standard schedule, Pass@1 without the bundle remains
low for both same-API and cross-API training
($1.9\%$ and $1.5\%$, respectively).
With extended training, same-API performance reaches $19.5\%$,
compared with $1.9\%$ for the cross-API control.
The paired gain is $17.6$ percentage points
($95\%$ CI $[14.7, 20.7]$), with a seed-averaged gain of
$17.9$ points.

This transfer remains limited relative to the $97.4\%$
pass rate on the same-API adapter's training tasks and the
$65.6\%$ achieved by the untuned model with the Full bundle.
With Full, same-API and cross-API training reach $74.6\%$
and $73.1\%$; their difference is not significant
($+1.5$ points, $95\%$ CI $[-1.9, +5.2]$).
\textbf{Training on the target API therefore improves performance
without documentation, while its additional benefit over
cross-API training is less significant when documentation is supplied.}

%% file: app/i_dataset_doc.tex
\section{Dataset Documentation and Licensing}
\label{app:dataset-doc}

We release the construction and evaluation code, candidate API
pool, tasks, knowledge bundles, model-specific task lists,
and experimental splits at
\url{https://github.com/MAPS-research/NovelAPIBench}.
Each task includes its description, context code, reference
completion, target API, difficulty label, and test harness.
Dataset statistics and construction procedures are provided in
Appendices~\ref{app:bench-stats} and~\ref{app:bench-pipeline};
adaptation splits are described in
Appendix~\ref{app:setup-rq3}.
Trained adapters, edited checkpoints, generation caches,
and raw evaluation logs are not included.

\paragraph{Sources and organization.}
The benchmark is constructed from public library source code,
docstrings, official documentation, and associated papers,
using GPT-5-mini for generation and sandboxed execution
for validation.
Tasks and knowledge bundles are organized as JSONL files.
Accompanying manifests record library versions, release boundaries,
filter settings, file checksums, and the task list for each
model-specific instance.
Excluded task identifiers and reasons are also documented.

\paragraph{Intended use.}
\system{} supports controlled studies of API knowledge,
retrieval, and parametric adaptation.
Its scope is single-target-API tasks in Python;
coverage and validation limitations are discussed in
Appendix~\ref{app:limitations}.

\paragraph{Licensing.}
\label{app:licenses}
Third-party source code, examples, and documentation excerpts
remain subject to their respective upstream licenses.
The release documentation records their provenance
and applicable licenses and include the corresponding license
texts and copyright notices.